%% file: arxiv.tex
\documentclass[runningheads]{llncs}

\usepackage{eccv}

\usepackage{eccvabbrv}

\usepackage{graphicx}
\usepackage{booktabs}
\usepackage{algorithm}
\usepackage{algorithmic}
\usepackage{multirow}
\usepackage{threeparttable}
\usepackage{amssymb}
\usepackage{xcolor}
\usepackage{pifont}% http://ctan.org/pkg/pifont
\newcommand{\cmark}{\ding{51}}%
\newcommand{\xmark}{\ding{55}}%
\usepackage{amsmath}
\usepackage{adjustbox}
\usepackage{float}
\usepackage{epigraph}
\definecolor{citecolor}{HTML}{0071bc}
\definecolor{ourscolor}{HTML}{c2d1e5}
\usepackage{colortbl}  % to use more color with table

\usepackage[misc]{ifsym}

\makeatletter
\def\blfootnote{\gdef\@thefnmark{}\@footnotetext}
\makeatother
\usepackage[accsupp]{axessibility}  % Improves PDF readability for those with disabilities.

\usepackage{hyperref}

\usepackage{orcidlink}

\begin{document}

\title{You Only Learn One Query: Learning Unified Human Query for Single-Stage Multi-Person Multi-Task Human-Centric Perception} 

\titlerunning{You Only Learn One Query}

%%%%%%%%% AUTHORS - PLEASE UPDATE
\author{Sheng Jin\inst{1,2}\thanks{Equal contribution. \quad \Letter~Corresponding authors.}\orcidlink{0000-0001-5736-7434} \and
Shuhuai Li\inst{2*}\orcidlink{0009-0006-0180-4248} \and
Tong Li\inst{2}\orcidlink{0009-0005-9993-2581} \and
Wentao Liu\inst{2}\orcidlink{0000-0001-6587-9878} \textsuperscript{\Letter} \and \\
Chen Qian\inst{2}\orcidlink{0000-0002-8761-5563} \and
Ping Luo\inst{1,3}\orcidlink{0000-0002-6685-7950} \textsuperscript{\Letter}
}

\authorrunning{S. Jin et al.}

\institute{$^{1}$ The University of Hong Kong \quad
$^{2}$ SenseTime Research and Tetras.AI \\
$^{3}$ Shanghai AI Laboratory \\
\email{js20@connect.hku.hk, lishuhuai@sensetime.com}}

\maketitle

\input{sec/0_abstract}

\input{sec/1_intro}

\input{sec/2_related_work}

\input{sec/3_dataset}
\input{sec/4_method}

\input{sec/5_experiment}

\input{sec/6_conclusion}

\bibliographystyle{splncs04}
\bibliography{egbib}

\clearpage
\appendix

\setcounter{table}{0}
\renewcommand{\thetable}{A\arabic{table}}
\setcounter{figure}{0}
\renewcommand{\thefigure}{A\arabic{figure}}

\input{sec/supp}

\end{document}

%% file: sec/0_abstract.tex
\begin{abstract}
Human-centric perception (\eg detection, segmentation, pose estimation, and attribute analysis) is a long-standing problem for computer vision. This paper introduces a unified and versatile framework (HQNet) for single-stage multi-person multi-task human-centric perception (HCP). 
Our approach centers on learning a unified human query representation, denoted as Human Query, which captures intricate instance-level features for individual persons and disentangles complex multi-person scenarios. Although different HCP tasks have been well-studied individually, single-stage multi-task learning of HCP tasks has not been fully exploited in the literature due to the absence of a comprehensive benchmark dataset. To address this gap, we propose COCO-UniHuman benchmark to enable model development and comprehensive evaluation. 
Experimental results demonstrate the proposed method's state-of-the-art performance among multi-task HCP models and its competitive performance compared to task-specific HCP models. Moreover, our experiments underscore Human Query's adaptability to new HCP tasks, thus demonstrating its robust generalization capability. Codes and data are available at \url{https://github.com/lishuhuai527/COCO-UniHuman}.

\keywords{Human-Centric Perception \and Unified Vision Model}
\end{abstract}

%% file: sec/1_intro.tex
\section{Introduction}

Human-centric perception (\eg pedestrian detection, 2D keypoint estimation, 3D mesh recovery, human segmentation and attribute recognition) have attracted increasing research attention owing to their widespread industrial applications such as sports analysis, virtual reality, and augmented reality. 

\begin{figure}[t]
\centering
    \includegraphics[width=0.6\textwidth]{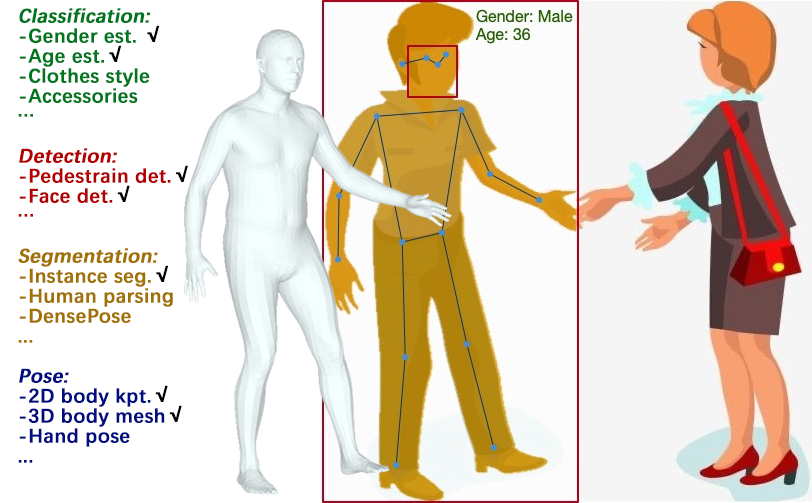}
\caption{Multi-person human-centric perception tasks can be categorized into 4 groups: classification, detection, segmentation and pose estimation. 
}
\label{fig:hcp}
\vspace{-5mm}
\end{figure}

The task of single-stage multi-person multi-task human-centric perception (HCP) has not been fully exploited in the literature due to the absence of a representative benchmark dataset. 
Consequently, previous studies~\cite{ci2023unihcp} resorted to training models on various datasets for each HCP task, which can introduce certain limitations.
Firstly, there is inherent scale variance across different datasets. For example, human detection datasets~\cite{lin2014microsoft} consist of scene images with multiple interacting people, while attribute recognition datasets~\cite{liu2017hydraplus} typically contain images with a single cropped person. This hampers the development of single-stage multi-task algorithms that can comprehensively address various HCP tasks as a unified problem.
Secondly, single-task datasets are often designed for specific application scenarios, resulting in strong dataset biases across different datasets. For example, some datasets~\cite{ionescu2013human3} are captured in controlled lab environments, while some~\cite{liu2017hydraplus} are captured from a surveillance viewpoint. Naively training models on a combination of these datasets inevitably introduces dataset biases and hinders performance in real-world, unconstrained scenarios.
Although there are separate benchmarks for individual HCP tasks, a comprehensive benchmark to simultaneously evaluate multiple HCP tasks is still lacking. 
To address this problem, we introduce a large-scale benchmark dataset called COCO-UniHuman, specifically designed for unified human-centric perceptions.
As shown in Figure~\ref{fig:hcp}, most popular HCP tasks can be grouped into four fundamental categories: classification, detection, segmentation, and pose estimation. The COCO-UniHuman dataset extends COCO dataset by extensively annotating gender and age labels for each person instance. It encompasses all these four categories, covering 7 diverse HCP tasks (marked with check marks in Figure~\ref{fig:hcp}).

Prior works on multi-person multi-task HCP have predominantly employed a multi-stage approach. These approaches typically involve employing a human detector to detect human instances, followed by task-specific models for each individual human perception task such as keypoint estimation and instance segmentation. However, these approaches exhibit three significant drawbacks.
Firstly, they suffer from the issue of early commitment: the performance of the whole pipeline highly relies on body detection, and there is no recourse to recovery if the body detector fails. 
Secondly, the run-time is proportional to the number of individuals present in the image, making them computationally expensive for real-time applications. In contrast, single-stage methods estimate all required properties for human-centric analysis in a single pass, resulting in improved efficiency.
Thirdly, these approaches overlook the potential inter-task synergy. Different HCP tasks are highly correlated as they share a common understanding of human body structure.
In this work, we develop a simple, straightforward and versatile baseline framework, called HQNet, for single-stage multi-task HCP. It unifies various distinct human-centric tasks, including pedetrian detection, human segmentation, 2D human keypoint estimation, 3D human mesh recovery, and human attribute analysis (specifically gender and age). 

Different HCP tasks have their own relevant features of diverse granularity to focus on. For instance, pedestrian detection emphasizes global semantic features; attribute recognition necessitates both global and local semantic cues; person segmentation relies on fine-grained semantic features; and pose estimation require fine-grained semantic and localization information. In this paper, we propose to learn unified all-in-one query representations, termed Human Query, to encode instance-specific features of diverse granularity from multiple perspectives.
Our work is inspired by DETR-based methods~\cite{carion2020end,zhang2022dino,liu2022dab,li2022dn,zhu2020deformable}, which employ learnable query embeddings to represent objects and infer the relations of the objects and the image features. 
This study expands upon these works by learning versatile instance-level query representations for general human-centric perceptions. 
In addition, we design HumanQuery-Instance Matching (HQ-Ins Matching) and Gender-aided human Model Selection (GaMS) mechanisms to further exploit the interactions among different HCP tasks and enhance the performance of multi-task HCP.

We highlight several noteworthy characteristics of HQNet.
(1) \textbf{Flexibility:} HQNet can readily integrate with diverse backbone networks, such as ResNet~\cite{he2016deep}, Swin~\cite{liu2021swin} and ViT~\cite{dosovitskiy2020image}. 
(2) \textbf{Scalibility:} the weight-sharing backbone, transformer encoder, and decoder in HQNet enables seamless integration with multiple tasks, with minimal overhead from each task-specific head, thus demonstrating remarkable scalability. 
(3) \textbf{Transferability:} Experiments demonstrate strong transferability of the learned Human Query to novel HCP tasks, such as face detection and multi-object tracking.

Our work makes the following key contributions:
(1) We introduce the COCO-UniHuman benchmark, a large-scale dataset that comprehensively covers all representative HCP tasks, \ie classification (gender and age estimation), detection (body and face detection), segmentation, and pose estimation (2D keypoint and 3D mesh recovery).
(2) We develop a simple yet effective baseline called HQNet, unifying multiple distinctive HCP tasks in a single-stage multi-task manner. The key idea is to learn unified all-in-one query representations, termed Human Query, which encode instance-specific features of diverse granularity from various perspectives. Additionally, we design HumanQuery-Instance (HQ-Ins) Matching and Gender-aided human Model Selection (GaMS) mechanisms to improve the performance of multi-task HCP.
(3) Our approach achieves state-of-the-art results on different HCP tasks, demonstrating the strong representation capability of the learnt Human Query. Furthermore, experiments show the strong transferability of the learned Human Query to novel HCP tasks, such as face detection and multi-object tracking.
We hope our work can shed light on future research on developing single-stage multi-person multi-task HCP algorithms.

%% file: sec/2_related_work.tex
\begin{table*}[t]
\caption{Overview of representative HCP datasets. ``\#Img'', ``\#Inst'', and ``\#ID'' mean the number of total images, instances and identities respectively.
``Crop'' indicates whether the images are cropped for ``face'' or ``body''.
* means head box annotation.
``group:n'' means age classification with n groups, ``real'' means real age estimation, and ``appa'' means apparent age estimation.}
\begin{center}
\vspace{-4mm}
\scalebox{0.72}{
\begin{tabular}{l|ccccccccccc}
    \hline
    Dataset & \#Img & \#Inst & \#ID & Crop  & BodyBox & FaceBox & BodyKpt & BodyMask & Gender & Age & Mesh \\ \hline
    \textit{Caltech}~\cite{dollar2009pedestrian} & 250K & 350K & 2.3K & \xmark &  \cmark & \xmark &  \xmark &  \xmark  & \xmark  & \xmark & \xmark   \\
    \textit{CityPersons}~\cite{zhang2017citypersons} & 5K & 32K & 32K  & \xmark &  \cmark & \xmark &  \xmark &  \xmark  & \xmark  & \xmark & \xmark  \\
    \textit{CrowdHuman}~\cite{shao2018crowdhuman} & 24K & 552K & 552K  & \xmark &  \cmark & *  &  \xmark &  \xmark  & \xmark  & \xmark  & \xmark  \\
    \hline
    \textit{MPII}~\cite{andriluka20142d} & 25K &  40K & -  & \xmark & \cmark &  * & \cmark & \xmark& \xmark  & \xmark & \xmark  \\
    \textit{PoseTrack}~\cite{andriluka2018posetrack} & 23K & 153K & - &  \xmark & \cmark &  * &  \cmark & \xmark  & \xmark  & \xmark & \xmark   \\ 
    \hline
    \textit{CIHP}~\cite{gong2018instance} & 38K & 129K & 129K  & \xmark &  \cmark   &  \xmark  & \xmark  &  \cmark  & \xmark  & \xmark & \xmark  \\ 
    \textit{MHP}~\cite{li2017multiple} & 5K & 15K & 15K  & \xmark &  \cmark   &  \xmark  & \xmark  &  \cmark  & \xmark  & \xmark  & \xmark  \\
    \hline
    \textit{CelebA}~\cite{liu2015faceattributes} & 200K &200K & 10K &  face &  \xmark &   \xmark &  \xmark   &  \xmark &   \cmark   &   group:4  & \xmark  \\
    \textit{APPA-REAL}~\cite{agustsson2017apparent} & 7.5K & 7.5K & 7.5K  & face &  \xmark &   \xmark &  \xmark  & \xmark &  \cmark  &  appa \& real & \xmark  \\
    \textit{MegaAge}~\cite{zhang2017quantifying} & 40K & 40K & 40K  & face &  \xmark &   \xmark &  \xmark  &  \xmark   &  \cmark  &  real & \xmark   \\
    \textit{WIDER-Attr}~\cite{li2016human} & 13K & 57K & 57K  & \xmark &  \cmark &  \xmark  &   \xmark  &  \xmark &   \cmark   &  group:6 & \xmark  \\
    \textit{PETA}~\cite{deng2014pedestrian} & 19K & 19K & 8.7K  & body & \xmark  &   \xmark &   \xmark  &  \xmark &   \cmark   &  group:4  & \xmark  \\
    \textit{PA-100K}~\cite{liu2017hydraplus} & 100K & 100K & -  & body &  \xmark &   \xmark &  \xmark  &  \xmark &   \cmark   & group:3 & \xmark  \\\hline
    \textit{OCHuman}~\cite{zhang2019pose2seg} & 5K & 13K & 13K  & \xmark &  \cmark  &   \xmark &   \cmark  &  \cmark  & \xmark  & \xmark & \xmark  \\
    \textit{COCO}~\cite{lin2014microsoft} & 200K & 273K & 273K & \xmark &  \cmark  & \xmark  &  \cmark  &  \cmark  &   \xmark& \xmark & \xmark \\
    \textit{COCO-WholeBody}~\cite{jin2020whole} & 200K & 273K & 273K & \xmark &  \cmark  &  \cmark  &  \cmark  &   \xmark & \xmark & \xmark  & \xmark  \\
    \hline
    \textit{COCO-UniHuman} & 200K & 273K & 273K & \xmark & \cmark  & \cmark & \cmark  &  \cmark & \cmark  & appa & \cmark  \\ 
    \hline       
\end{tabular}
}
\end{center}
\vspace{-4mm}
\label{tab:dataset}
\end{table*}

\section{Related Works}

\subsection{Human-Centric Perception Tasks and Datasets} 
Approaches to multi-person human-centric perception (HCP) can be categorized into top-down, bottom-up, and single-stage methods.
\textbf{Top-down methods} follow a detect-then-analyze approach. They first localize human instances, and then perform single person analysis. Top-down approaches can be divided into two types: those using separate pre-trained detectors and task-specific perception models~\cite{kanazawa2018end,zeng20203d,lin2021end,sun2019deep,wang2020deep,jiang2022posetrans,xu2021vipnas}, and those jointly learning detection and perception modules~\cite{he2017mask,alp2018densepose}.
\textbf{Bottom-up methods} learn instance-agnostic keypoints/masks and cluster them using integer linear programming~\cite{insafutdinov2016deepercut,kirillov2017instancecut,jin2019multi}, heuristic greedy parsing~\cite{cao2017realtime,papandreou2018personlab}, embedding clustering~\cite{newell2017associative,kong2018recurrent}, or learnable clustering~\cite{jin2020differentiable}.
\textbf{Single-stage methods} directly predict keypoints or masks for each individual, with different representations for 2D keypoint estimation (coordinate-based~\cite{tian2019directpose,nie2019single,wei2020point,xue2022learning}, heatmap-based~\cite{wang2022contextual,shi2021inspose}, or hybrid~\cite{zhou2019objects,mao2021fcpose,geng2021bottom}), 3D mesh recovery~\cite{sun2021monocular,lin2023one} and segmentation (contour-based~\cite{xie2020polarmask} or mask-based~\cite{bolya2019yolact}).
While existing approaches focus on individual HCP tasks, we aim to unify HCP by learning a single model that handles multiple tasks simultaneously, enabling a comprehensive understanding of humans.
As shown in Table~\ref{tab:dataset}, there are task-specific datasets separately annotated for different HCP tasks, including pedestrian detection~\cite{zhang2024when,dollar2009pedestrian,zhang2017citypersons,shao2018crowdhuman}, keypoint estimation~\cite{andriluka20142d,andriluka2018posetrack}, segmentation~\cite{gong2018instance,li2017multiple}, and attribute recognition~\cite{zhang2017quantifying,liu2017hydraplus}. Datasets for multiple HCP tasks also exist. COCO~\cite{lin2014microsoft} offers thorough annotations: body box, keypoints, and segmentation mask. COCO-WholeBody~\cite{jin2020whole,xu2022zoomnas} provides dense annotations of face/hand boxes and 133 whole-body keypoints. Our COCO-UniHuman dataset further extends COCO-WholeBody featuring extensive gender, age and mesh annotations.

\subsection{Unified Methods for HCP}
\textbf{General network architecture for different HCP tasks.}
Some works design general network backbones, including CNN-based~\cite{wang2020deep} and Transformer-based backbones~\cite{zeng2022not}. Others unify HCP tasks with novel perception heads, such as UniHead~\cite{liang2022unifying} and UniFS~\cite{jin2024unifs}.
Unlike these methods, which employ separate task-specific models, we consolidate diverse HCP tasks within a single network.
\textbf{Pre-training on HCP tasks.} There are also works~\cite{hong2022versatile,chen2023beyond,tang2023humanbench} on pre-training on diverse human-centric tasks with large-scale data. 
More recently, UniHCP~\cite{ci2023unihcp} presents a unified vision transformer model to perform multitask pre-training at scale.  It employs task-specific queries for attending to relevant features, but tackles one task at a time. Unlike ours,  our approach simultaneously solves multiple HCP tasks in a single forward pass. 
Our approach contrasts with these pre-training based methods by avoiding pre-training, minimizing fine-tuning, and circumventing resource-intensive multi-dataset training. Unlike them, we handle multiple HCP tasks concurrently in a single-stage, multi-task manner, diverging from their single-person focus.
\textbf{Co-learning on HCP tasks.} Many works have investigated the correlations between pairs of HCP tasks~\cite{zhang2014panda,tian2015pedestrian,lin2022fp,nie2018mutual,nie2018human}. 
We propose a single-stage model that learns a general unified representation to handle all representative human-centric perception tasks simultaneously.

\subsection{Object-Centric Representation Learning}
DETR~\cite{carion2020end} pioneers learnable object queries to represent objects and interact with image features. Deformable DETR~\cite{zhu2020deformable} introduces deformable attention modules to focus on key sampling points, enhancing convergence speed. DAB-DETR~\cite{liu2022dab} treats each positional query as a dynamic 4D anchor box, updated across decoder layers. DN-DETR~\cite{li2022dn} employs denoising training for faster convergence. Recently, DINO~\cite{zhang2022dino} amalgamates these techniques, introducing a mixed query selection and look-forward-twice strategy to expedite and stabilize training.
Our work is inspired by DETR-based methods. Especially we build upon DINO and extend it to develop a versatile framework for single-stage multi-task HCP, unifying multiple distinct human-centric tasks.

%% file: sec/3_dataset.tex
\section{COCO-UniHuman Dataset}

COCO-UniHuman v1 dataset is the first large-scale dataset, which provides annotations for all four representative HCP tasks in multi-person scenarios. 
Building upon COCO~\cite{lin2014microsoft,joo2021exemplar} dataset, we have enriched the annotations by including gender, age, 3D body mesh information for each individual.

\subsection{Data Annotation}
\subsubsection{Human Attribute Annotation.}
To ensure accurate annotations, we employ trained annotators to manually label the gender and apparent age for each human instance in the dataset. 
We discard images full of non-human objects, and exclude all \textit{Small} category persons that are hardly attribute-recognizable. 
\textbf{Gender annotation.} For each valid human instance, we adopt a body-based annotation approach. Using the provided human bounding boxes, we crop the body images and request annotators to label the gender. To maintain data quality, we conduct quality inspections and manual corrections throughout the labeling process.
\textbf{Age annotation.}
To enhance the quality of annotation, we employ a two-stage strategy based on body-based annotation. Similar to gender annotation, age annotation is also performed on cropped body images. We implement a coarse-to-fine two-stage annotation strategy, considering age group annotation to be comparatively easier than apparent age annotation~\cite{agustsson2017apparent}.
In the first stage, age groups are annotated. Following~\cite{li2016human}, we divide the age ranges into six groups, \ie ``baby'', ``kid'', ``teen'', ``young'', ``middle aged'', and ``elderly''.
For each cropped person image, we request a group of 10 annotators to independently and repeatedly label the age groups (6-category classification task). We take the mode of the 10 votes as the ground-truth age group. 
In the second stage, the apparent age is annotated. Given the age group as a prior, a group of 10 annotators independently and repeatedly annotate the apparent age. Consequently, we obtain 10 votes for each human instance. We remove the outliers and take the average as the final ground-truth apparent age. As a summary, the dataset contains over 1M apparent votes.
Experiments validate the effectiveness of the body-based annotation strategy and the two-stage annotation strategy (see Supplementary). 
\textbf{Mesh Annotation.}
We follow~\cite{joo2021exemplar} to apply Exemplar Fine-Tuning (EFT) method with the gender neutral model to generate 3D pseudo-ground truth SMPL parameters. To ensure data quality, we only retain instances with at least 12 keypoint annotations where all limbs are visible and filter out low-quality data manually.

\subsection{Data Uniqueness} 
The newly introduced dataset possesses several noteworthy properties in comparison to existing HCP datasets.
\textbf{(1) Comprehensiveness:} This is the first large-scale multi-person HCP dataset that encompasses all four basic HCP tasks, \ie classification, detection, segmentation, keypoint localization, and 3D body mesh recovery in multi-person scenarios. It facilitates the development and evaluation of single-stage multi-person multi-task HCP algorithms. 
\textbf{(2) Large scale and high diversity:} With over 200,000 images and 273,000 identities, this dataset exhibits significant variations in terms of lighting conditions, image resolutions, human poses, and indoor/outdoor environments.
\textbf{(3) Multi-person attribute recognition:}
Unlike most existing human attribute recognition datasets that solely provide single-person center cropped images, our proposed dataset offers a valuable benchmark for multi-person attribute recognition in challenging scenarios.
\textbf{(4) Body-based apparent age estimation:}
While previous research has primarily focused on predicting a person's age based on facial images, our dataset emphasizes the utilization of richer visual cues derived from whole-body images. Incorporating body-based visual cues such as skin elasticity, body posture, and body height proves beneficial for estimating a person's age, particularly in situations where the facial image lacks clarity (\eg captured from a distance). Notably, existing large-scale pedestrian attribute datasets~\cite{deng2014pedestrian} typically only offer coarse age group annotations, while facial attribute datasets~\cite{agustsson2017apparent} often provide fine-grained apparent or real age annotations. Our proposed dataset bridges this gap and serves as the pioneering large-scale dataset for body-based apparent age estimation in the wild. 
\textbf{(5) Enhanced human representation:} The extended human attribute labels and 3D human mesh information provide additional descriptive information about individuals beyond the existing labels. By leveraging these information, models can learn improved representations of humans, consequently enhancing the performance of other HCP tasks. 

%% file: sec/4_method.tex
\begin{figure*}[t]
\centering
    \includegraphics[width=0.9\textwidth]{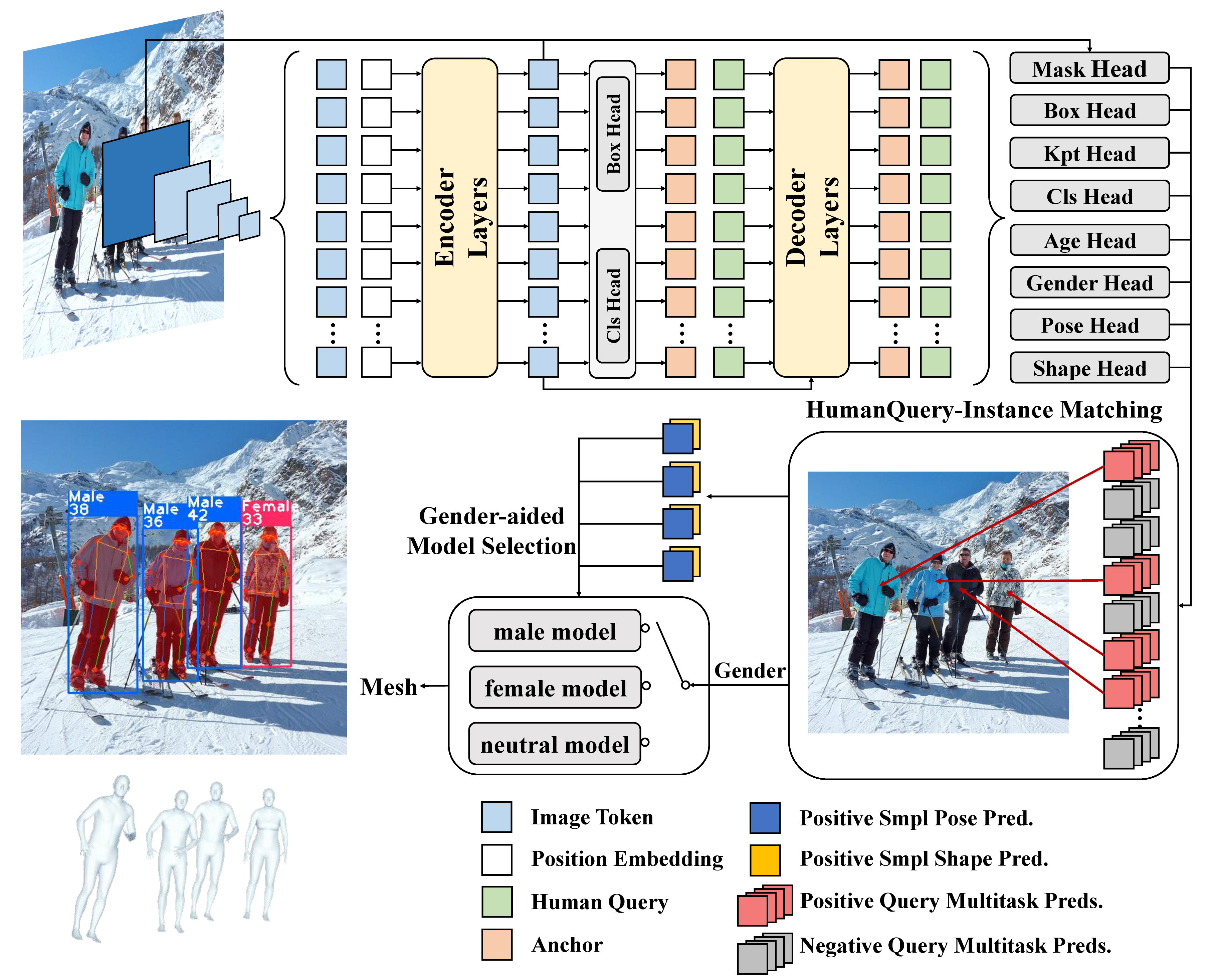}
\caption{Overview of HQNet. 
HQNet unifies various representative HCP tasks in a single network by learning shared Human Query. 
}
\label{fig:overview}
\end{figure*}

\section{Method}

\subsection{Overview}

This study endeavors to develop a single-stage framework that supports a wide range of human-centric perception (HCP) tasks. The key is to learn a comprehensive human representation, which can be universally employed across various HCP tasks. To achieve this, we employ a query-based methodology and investigate the feasibility of representing each human instance as a single shared query.
Unlike previous task-specific HCP models that may incorporate specialized designs tailored to specific tasks (\eg ``mask-enhanced anchor box initialization'' in Mask DINO~\cite{li2023mask}), our approach aims to handle various human-centric analysis tasks in a unified manner. To maximize knowledge sharing among various HCP tasks, we attempt to share most weights across different HCP tasks. 

As illustrated in Figure~\ref{fig:overview}, our framework consists of four key components: a backbone network, a Transformer encoder, a task-shared Transformer decoder and task-specific heads. The backbone network, such as ResNet~\cite{he2016deep}, takes an image as input and produces multi-scale features. These features, along with corresponding positional embeddings, are then passed through the Transformer encoder to enhance the feature representation.
We use the mixed query selection technique to select initial anchor boxes as positional queries for the Transformer decoder. Following DINO~\cite{zhang2022dino}, we only initialize the positional queries but do not initialize content queries.
Unlike previous approaches that employ task-specific Transformer decoders, we propose to use a task-shared decoder for all HCP tasks. The Transformer decoder incorporates the deformable attention~\cite{zhu2020deformable} to refine the queries across decoder layers. 
We refer to the refined content queries as ``Human Query'' as they encode diverse information pertaining to human instances. Finally, the Human Queries are fed into each light-weight task-specific head for final prediction. 

\subsection{Task-Shared Transformer Decoder}
Queries in DETR-like models are formed by two parts: positional queries and content queries. Each positional query is formulated as a 4D anchor box, encoding the center x-y coordinates, width and height of the box, respectively. Our content query, denoted as Human Query, encapsulates various features (local and global appearance features, as well as coarse- and fine-grained localization features) specific to each instance. 
To enhance training stability and acceleration, we employ Contrastive DeNoising (CDN) as introduced in DINO~\cite{zhang2022dino}. Notably, we observe that incorporating auxiliary DeNoise losses for other tasks (\eg segmentation and pose) does not yield significant improvements. Consequently, we only apply DN losses for human detection.

\textbf{HumanQuery-Instance Matching.}
To ensure consistent and unique predictions for each ground-truth instance across all HCP tasks, \ie classification (Cls.), detection (Det.), pose (Pose.), and segmentation (Seg.), we employ HumanQuery-Instance (HQ-Ins) Matching. $\lambda_{cls}L_{cls} + \lambda_{det}L_{det} + \lambda_{seg}L_{seg} + \lambda_{pose}L_{pose},$ where $\lambda$ are loss weights. Details can be found in Supplementary.

\subsection{Task-Specific Heads}
To ensure scalability, we categorize HCP tasks into three groups and design specific implementation paradigms for each category. 
\textbf{Coordinate prediction} tasks (\eg object detection and keypoint estimation) share common reference points with bounding box prediction and directly regress the normalized offsets of each point.
\textbf{Dense prediction} tasks (\eg instance segmentation and human parsing) follow the design of Mask DINO~\cite{li2023mask}, which involves constructing a high-resolution pixel embedding map by integrating features from both the backbone and the Transformer encoder. By performing a dot-product operation between the content query embedding and the pixel embedding map, an instance-aware pixel embedding map is generated, facilitating pixel-level classification.
\textbf{Classification} tasks (\eg determining if an instance is human, gender and age estimation) directly map the Human Query to the classification prediction results, as the Human Query inherently encodes the positional information.

To minimize the overhead of incorporating new tasks, we employ lightweight task-specific heads.
\textbf{Human detection head.} A 3-layer multi-layer perceptron (MLP) with a hidden dimension of $d$ is utilized to predict the normalized center x-y coordinates, height, and width of the bounding box w.r.t. the input image. Additionally, a linear projection layer (FC) is employed to predict the class label (human or non-human).
\textbf{2D keypoint estimation head.} Following the coordinate prediction paradigm, the learned Human Query is fed into a pose regression head (MLP) to regress the relative pose offsets w.r.t. the shared reference points of the detection head. A confidence prediction head (FC) is used to predict confidence score of having visible keypoints. Following PETR~\cite{shi2022end}, joint decoder layers are employed to refine body poses by leveraging structured relations between body keypoints. An auxiliary heatmap branch is used to aid training and discarded during testing.
\textbf{Human instance segmentation head.}
A 3-layer MLP is used to process the instance-aware pixel embedding map and output a one-channel mask, which is then upsampled to match the original input image size.
\textbf{Human attribute head.} The gender estimation head and the age estimation head operate in parallel. Both heads consist of two-layer MLPs. Gender estimation involves binary classification, while age estimation is formulated as an 85-class ([1, 85]) classification with softmax expected value~\cite{rothe2015dex} estimation.
\textbf{3D mesh recovery head.} Two 3-layer MLPs with the same hidden dimension are used to predict the pose and shape parameters respectively. These parameters are then fed into a SMPL body model to generate the 3D body meshes.

\subsubsection{Gender-aided human Model Selection (GaMS).} There are three versions of SMPL models: male, female and neutral. Previous works usually use the neutral model because of the lack of gender annotation. COCO-UniHuman has both gender and 3D mesh annotations. To improve the performance of 3D mesh, we employ Gender-aided human Model Selection (GaMS) which selects different SMPL models by gender labels during the training and the inference stage.

%% file: sec/5_experiment.tex
\section{Experiments}

\begin{table*}[htb]
\caption{\textbf{Comparisons with task-specific and multi-task models} on the COCO-UniHuman \texttt{val} set. We report AP for the ``Person'' category without \textit{Small} category person.
* denotes models trained to handle general 80 classes. 
$\dag$ denotes flip testing.
We compare with $\diamondsuit$ top-down, $\heartsuit$ bottom-up, $\bigstar$ one-stage approaches.
}
\centering
\small
\scalebox{0.71}{
\begin{tabular}{l c c c c c c c c c c c c c c c c}
\toprule
\multirow{2}{*}{Model} & \multirow{2}{*}{Backbone} & \multicolumn{3}{c}{Det.} &  \multicolumn{3}{c}{Seg.} & \multicolumn{3}{c}{Pose (Kpt.)}  & \multicolumn{3}{c}{Cls. (Gender)}  & \multicolumn{3}{c}{Cls. (Age)} \\
\cmidrule(r){3-5}
\cmidrule(r){6-8}
\cmidrule(r){9-11}
\cmidrule(r){12-14}
\cmidrule(r){15-17}
& & $\text{AP}$ & $\text{AP}^{M}$ & $\text{AP}^{L}$ &  $\text{AP}$  & $\text{AP}^{M}$ & $\text{AP}^{L}$ &  $\text{AP}$  & $\text{AP}^{M}$ & $\text{AP}^{L}$ &  $\text{AP}$  & $\text{AP}^{M}$ & $\text{AP}^{L}$  &  $\text{AP}$  & $\text{AP}^{M}$ & $\text{AP}^{L}$\\
\toprule
Faster R-CNN~\cite{renNIPS15fasterrcnn}  & R-50 & 65.3 & 61.5 & 71.2 & \xmark & \xmark & \xmark & \xmark & \xmark & \xmark & \xmark & \xmark & \xmark &  \xmark  & \xmark & \xmark  \\ 
IterDETR~\cite{zheng2022progressive} & R-50 & 71.8 & 66.0 & 78.9 & \xmark & \xmark & \xmark & \xmark & \xmark & \xmark & \xmark & \xmark & \xmark &  \xmark  & \xmark & \xmark \\
DINO~\cite{zhang2022dino} & R-50 & 73.3 & 68.1 & 79.9 & \xmark & \xmark & \xmark & \xmark & \xmark & \xmark & \xmark & \xmark & \xmark &  \xmark  & \xmark & \xmark \\
\cmidrule(lr){1-17}
$\diamondsuit$ Mask R-CNN~\cite{he2017mask} & R-50-FPN &  66.7 &  62.3 & 73.1 & 58.4 & 51.8 & 66.2 & \xmark  & \xmark & \xmark & \xmark  & \xmark & \xmark &  \xmark  & \xmark & \xmark  \\
$\bigstar$ PolarMask~\cite{xie2020polarmask} & R-50-FPN &  \xmark  & \xmark & \xmark  & 45.1 & 38.5 & 57.1 & \xmark  & \xmark & \xmark & \xmark  & \xmark & \xmark &  \xmark  & \xmark & \xmark  \\ 
$\bigstar$ YOLACT~\cite{bolya2019yolact} & R-50-FPN &  \xmark  & \xmark & \xmark  & 47.4 & 40.1 & 61.2 & \xmark  & \xmark & \xmark  & \xmark  & \xmark & \xmark &  \xmark  & \xmark & \xmark \\ 
$\bigstar$ MEInst~\cite{zhang2020mask} & R-50-FPN &  \xmark  & \xmark & \xmark  & 49.3 & 42.3 & 57.6 & \xmark  & \xmark & \xmark  & \xmark  & \xmark & \xmark &  \xmark  & \xmark & \xmark  \\ 
$\bigstar$ CondInst~\cite{tian2020conditional} & R-50-FPN &  \xmark  & \xmark & \xmark & 54.8 & 43.3 & 69.0 & \xmark  & \xmark & \xmark  & \xmark  & \xmark & \xmark &  \xmark  & \xmark & \xmark \\ 
$\bigstar$ Mask DINO~\cite{li2023mask} & R-50  & 72.3 & 66.5 & 79.5 & 64.8 & 57.3 & 73.4 & \xmark  & \xmark & \xmark  & \xmark  & \xmark & \xmark  & \xmark & \xmark & \xmark \\
\cmidrule(lr){1-17}
$\diamondsuit$ SBL$^\dag$~\cite{xiao2018simple} & R-50 & \xmark & \xmark & \xmark  & \xmark & \xmark & \xmark & 70.4 & 67.1 & 77.2 & \xmark & \xmark & \xmark &  \xmark  & \xmark & \xmark  \\
$\diamondsuit$ Swin$^\dag$~\cite{liu2021swin} & Swin-L & \xmark & \xmark & \xmark  & \xmark & \xmark & \xmark & 74.3 &  70.6 & 81.2 & \xmark & \xmark & \xmark &  \xmark  & \xmark & \xmark \\
$\diamondsuit$ HRNet$^\dag$~\cite{sun2019deep} & HRNet-32 & \xmark & \xmark & \xmark & \xmark & \xmark & \xmark & 74.4 &  70.8 & 81.0 & \xmark & \xmark & \xmark &  \xmark  & \xmark & \xmark \\
$\diamondsuit$ ViTPose$^\dag$~\cite{xu2022vitpose} & ViT-L & \xmark & \xmark & \xmark & \xmark & \xmark & \xmark & 78.2 & 74.5 & 85.4 & \xmark & \xmark & \xmark &  \xmark  & \xmark & \xmark \\
$\diamondsuit$ PRTR$^\dag$~\cite{li2021pose} & R-50 & \xmark & \xmark & \xmark  & \xmark & \xmark & \xmark & 68.2 & 63.2 & 76.2 &\xmark & \xmark & \xmark &  \xmark  & \xmark & \xmark \\
$\heartsuit$ HrHRNet$^\dag$~\cite{cheng2020higherhrnet} & HRNet-w32 & \xmark & \xmark & \xmark & \xmark & \xmark & \xmark & 67.1 & 61.5 & 76.1  & \xmark & \xmark & \xmark &  \xmark  & \xmark & \xmark  \\ 
$\heartsuit$ DEKR$^\dag$~\cite{geng2021bottom} & HRNet-w32 & \xmark & \xmark & \xmark  & \xmark & \xmark & \xmark & 68.0 & 62.1 & 77.7 & \xmark & \xmark & \xmark &  \xmark  & \xmark & \xmark \\  
$\heartsuit$ SWAHR$^\dag$~\cite{luo2021rethinking} & HRNet-w32 & \xmark & \xmark & \xmark  & \xmark & \xmark & \xmark & 68.9 & 63.0 & 77.4 & \xmark & \xmark & \xmark &  \xmark  & \xmark & \xmark  \\ 
$\bigstar$ CID$^\dag$~\cite{wang2022contextual} & R-50-FPN & \xmark & \xmark & \xmark  & \xmark & \xmark & \xmark & 52.0 & 48.6 & 58.0 & \xmark & \xmark & \xmark &  \xmark  & \xmark & \xmark\\ 
$\bigstar$ CID$^\dag$~\cite{wang2022contextual} & HRNet-w32 & \xmark & \xmark & \xmark  & \xmark & \xmark & \xmark  & 69.8 & 64.0 & 78.9 & \xmark & \xmark & \xmark &  \xmark  & \xmark & \xmark \\ 
$\bigstar$ FCPose~\cite{mao2021fcpose} & R-50 & \xmark & \xmark & \xmark & \xmark & \xmark & \xmark & 63.0 & 59.1 & 70.3 & \xmark & \xmark & \xmark &  \xmark  & \xmark & \xmark \\ 
$\bigstar$ InsPose~\cite{shi2021inspose} & R-50 & \xmark & \xmark & \xmark & \xmark & \xmark & \xmark & 65.2 & 60.6 & 72.2 & \xmark & \xmark & \xmark &  \xmark  & \xmark & \xmark \\  
$\bigstar$ PETR~\cite{shi2022end} & R-50 & \xmark & \xmark & \xmark & \xmark & \xmark & \xmark & 68.8 & 62.7 & 77.7 & \xmark & \xmark & \xmark &  \xmark  & \xmark & \xmark \\  
\cmidrule(lr){1-17}
$\diamondsuit$ StrongBL~\cite{jia2020rethinking} & R-50 & - & - & - & \xmark & \xmark & \xmark & \xmark & \xmark & \xmark & 46.4 & 35.2 & 53.2 &  \xmark  & \xmark & \xmark \\
$\diamondsuit$ Mask R-CNN~\cite{he2017mask} & R-50 &  66.3 & 61.9  & 72.8  & \xmark & \xmark & \xmark & \xmark & \xmark & \xmark & 46.7 & 36.3 & 52.8 &  \xmark  & \xmark & \xmark \\
\cmidrule(lr){1-17}
$\diamondsuit$ StrongBL~\cite{jia2020rethinking} & R-50 & - & - & - & \xmark & \xmark & \xmark & \xmark & \xmark & \xmark & \xmark &  \xmark  & \xmark & 42.3 & 31.9 & 48.3 \\
$\diamondsuit$ Mask R-CNN~\cite{he2017mask} & R-50 & 66.3 &  62.1 & 72.5  & \xmark & \xmark & \xmark & \xmark & \xmark & \xmark & \xmark &  \xmark  & \xmark & 37.4 & 27.9 & 43.3 \\
\cmidrule(lr){1-17}
$\diamondsuit$ Pose2Seg~\cite{zhang2019pose2seg} & R-50-FPN & \xmark & \xmark & \xmark & 55.5 &  49.8  & 67.0 &  59.9 & - & -   & \xmark & \xmark & \xmark &  \xmark  & \xmark & \xmark   \\  
$\heartsuit$ MultiPoseNet~\cite{abdulnabi2015multi} & R-50 & -  & 58.0 & 68.1 & - & - & - & 62.3 & 57.7 & 70.4 & \xmark & \xmark & \xmark  &  \xmark  & \xmark & \xmark \\
$\heartsuit$ PersonLab~\cite{papandreou2018personlab} & R-152  & \xmark & \xmark & \xmark & - & 48.3  & 59.5 & 66.5 &  62.3  & 73.2  & \xmark & \xmark & \xmark &  \xmark  & \xmark & \xmark \\
$\bigstar$ CenterNet~\cite{zhou2019objects} & Hourglass & - & -  & - & \xmark & \xmark & \xmark & 64.0 & 59.4 & 72.1 & \xmark & \xmark & \xmark &  \xmark  & \xmark & \xmark \\
$\bigstar$ LSNet-5~\cite{zhang2021location} & DLA-34 & \xmark  & \xmark & \xmark & 56.2 & 44.2 & 71.0  & -  & - & - & \xmark  & \xmark & \xmark &  \xmark  & \xmark & \xmark  \\
$\bigstar$ UniHead$^*$~\cite{liang2022unifying} & R-50-FPN &  67.3 & 62.6 & 74.4  & 38.6 & 37.2 & 42.2 & 57.5 & 55.3 & 61.9 & \xmark & \xmark & \xmark &  \xmark  & \xmark & \xmark  \\
\cmidrule(lr){1-17}
\rowcolor{ourscolor}
$\bigstar$ HQNet (D+S) & R-50   & 73.0 & 68.0 & 79.4  & 63.6 & 57.6 & 72.1 & \xmark & \xmark & \xmark & \xmark & \xmark & \xmark & \xmark & \xmark & \xmark \\
\rowcolor{ourscolor}
$\bigstar$ HQNet (D+S+P) & R-50   & 74.5 & 70.3 & 80.1  & 65.7  & 58.7 & 73.8 & 69.5 & 64.4 & 77.0 & \xmark & \xmark & \xmark & \xmark & \xmark & \xmark  \\
\rowcolor{ourscolor}
$\bigstar$ HQNet (D+S+P+C) & R-50   & 74.9  & 70.4 & 80.7  &  65.8 & 58.7 &  73.9 & 69.3 & 63.8 & 77.3 &  56.0 & 42.5 & 63.3 & 53.8 & 39.7 & 61.2  \\
\midrule
\rowcolor{ourscolor}
$\bigstar$ HQNet & Swin-L &  77.3 & 73.3 & 82.7  & 68.1 & 60.9 & 75.9 & 72.6 & 67.4 & 80.1  & 57.9 & 43.1 & 65.8 & 56.2 & 41.5 & 63.9 \\
\rowcolor{ourscolor}
$\bigstar$ HQNet & ViT-L &  78.0 & 73.6 & 83.7  & 68.6 & 61.4 & 76.5  & 75.3 & 69.8 & 83.5 & 58.0 & 44.7 & 65.0 & 58.0 & 40.9 & 66.7 \\
\bottomrule
\end{tabular}
}
\label{tab:coco}
\vspace{-5mm}
\end{table*}

\subsection{Dataset and Evaluation Metric}

\textbf{COCO-UniHuman Dataset.} 
Our model training exclusively employs COCO-UniHuman train data (in addition to ImageNet pre-training).  We follow DINO~\cite{zhang2022dino} for augmentation and adopt the 100-epoch training schedule.
Model evaluation takes place on COCO-UniHuman \texttt{val} set (2693 images). Due to limitations of the COCO \texttt{test-dev} evaluation server, which lacks support for ``Person'' category evaluation and attribute recognition, we mainly report results on the \texttt{val} set. 
The evaluation is based on the standard COCO metrics including Average Precision (AP), $\text{AP}^{M}$ for medium-sized persons and $\text{AP}^{L}$ for large-sized persons. Following~\cite{zhang2019pose2seg,zhang2021location}, we exclude the \emph{Small} category persons during evaluation due to the lack of annotations in COCO.
For attribute recognition, we also use AP with Age-10 metric for evaluation, where the age estimation is considered correct if the prediction error is no larger than 10. 
For human mesh recovery, we evaluate pose accuracy using MPJPE
(Mean Per Joint Position Error) w.r.t. root relative poses and PA-MPJPE (Procrustes-Aligned MPJPE), which is MPJPE calculated after rigid alignment of predicted pose with the ground truth.

\begin{table}[tb]
\caption{Results of 3D human mesh recovery on the COCO-UniHuman \texttt{val} set. HQNets are jointly trained with all HCP tasks (D+S+P+C). $\downarrow$ means lower is better.
}
\centering
\small
\vspace{-1mm}
\scalebox{0.79}{
    \setlength{\tabcolsep}{10pt}
    \begin{tabular}{l c c c c}
    \toprule
    \multirow{2}{*}{Model} & \multirow{2}{*}{Backbone} & \multirow{2}{*}{Bbox} &  \multicolumn{2}{c}{Pose (Mesh.)} \\
       &  & & $\text{MPJPE}\downarrow$ & $\text{PA-MPJPE}\downarrow$ \\
    \midrule
    $\diamondsuit$ HMR~\cite{kanazawa2018end} & R-50 & GT & 109.62 & 72.03  \\
    $\diamondsuit$ HMR+~\cite{pang2022benchmarking} & R-50 & GT & 78.06 & 50.36  \\
    $\bigstar$ ROMP~\cite{sun2021monocular} & R-50 & - & 119.52 & 72.27 \\
    \midrule
    $\bigstar$ HQNet w/o GaSM & R-50   & - & 87.00  &  54.92\\
    $\bigstar$ HQNet  & R-50  & - & 84.74  & 50.80 \\
    \midrule
    $\bigstar$ HQNet & ViT-L & - & 76.31  & 48.26  \\
    \bottomrule
    \end{tabular}
}
\vspace{-4mm}
\label{tab:mesh}
\end{table}

\textbf{OCHuman Dataset}~\cite{zhang2019pose2seg} is a large benchmark that focuses on heavily occluded humans. It contains no training samples and is intended solely for evaluation purposes. Following~\cite{zhang2019pose2seg}, we train models on the COCO \texttt{train} set and evaluate models on OCHuman \texttt{val} set (4731 images) and \texttt{test} set (8110 images). 

\subsection{Results on COCO-UniHuman Dataset}
We compare our method to task-specific and multi-task HCP models on the COCO-UniHuman dataset in Table~\ref{tab:coco} and Table~\ref{tab:mesh}. Our models outperform multi-task HCP models and achieves very competitive results against task-specific HCP models. Details about the baselines can be found in Supplementary. ``D'', ``S'', ``P'', ``C'' mean model training with Detection (Det.), Segmentation (Seg.), Pose and Classification (Cls.) task respectively. 

\textbf{Comparison with task-specific HCP models.} For human detection, we compare three baseline approaches, \ie Faster-RCNN~\cite{renNIPS15fasterrcnn}, IterDETR~\cite{zheng2022progressive} and DINO~\cite{zhang2022dino}. 
For human instance segmentation, we contrast HQNet with state-of-the-art general and human-specific instance segmentation methods, including Mask R-CNN~\cite{he2017mask}, PolarMask~\cite{xie2020polarmask}, MEInst~\cite{zhang2020mask}, YOLACT~\cite{bolya2019yolact}, and CondInst~\cite{tian2020conditional}. 
For human pose estimation, we compare with several representative top-down methods (SBL~\cite{xiao2018simple}, HRNet~\cite{sun2019deep}, Swin~\cite{liu2021swin}, ViTPose~\cite{xu2022vitpose} and PRTR~\cite{li2021pose}), bottom-up approaches (HrHRNet~\cite{cheng2020higherhrnet}, DEKR~\cite{geng2021bottom}, and SWAHR~\cite{luo2021rethinking}) and single-stage approaches (FCPose~\cite{mao2021fcpose}, InsPose~\cite{shi2021inspose}, PETR~\cite{shi2022end}
and CID~\cite{wang2022contextual}). 
For gender and age estimation, we establish baselines using StrongBL~\cite{jia2020rethinking} and Mask R-CNN~\cite{he2017mask}. For mesh, we compare with HMR~\cite{kanazawa2018end}, HMR+~\cite{pang2022benchmarking} and ROMP~\cite{sun2021monocular}.
Our approach achieves very competitive performance compared to other task-specific HCP models when using the R-50 backbone. Moreover, with stronger backbones such as Swin-L and ViT-L, we achieve SOTA among single-stage approaches.

\begin{table}[tb]
\caption{Comparison with state-of-the-art models on the OCHuman dataset. 
$\dag$ denotes flip testing.
We compare with $\diamondsuit$ top-down, $\heartsuit$ bottom-up, $\bigstar$ one-stage approaches.
}
\centering
\small
\scalebox{0.73}{
\setlength{\tabcolsep}{10pt}
\begin{tabular}{lc ccc | ccc}
\toprule
\multirow{3}{*}{Model} & \multirow{3}{*}{Backbone} & \multicolumn{3}{c}{\textbf{OCHuman Val}} &  \multicolumn{3}{c}{\textbf{OCHuman Test}} \\
& &  Det. & Seg. & Pose (Kpt.) &  Det. & Seg. & Pose (Kpt.) \\
\toprule
$\diamondsuit$ Mask R-CNN~\cite{he2017mask} & R-50-FPN & - & 16.3 & \xmark  & - & 16.9 & \xmark  \\
$\diamondsuit$ SBL$^\dag$~\cite{xiao2018simple} &  R-50  & \xmark  &   \xmark  & 37.8  & \xmark & \xmark &  30.4 \\  %from HGG paper
$\diamondsuit$ Pose2Seg~\cite{zhang2019pose2seg} & R-50-FPN & \xmark & 22.2 & 28.5 & \xmark & 23.8 & 30.3 \\
$\heartsuit$ AE$^\dag$~\cite{newell2017associative} &  Hourglass & \xmark &  \xmark  & 32.1   & \xmark &  \xmark  &  29.5   \\  %from HGG paper
$\heartsuit$ HGG$^\dag$~\cite{jin2020differentiable} & Hourglass  & \xmark  &   \xmark & 35.6  &  \xmark & \xmark  &  34.8  \\  %from HGG paper
$\heartsuit$ DEKR$^\dag$~\cite{geng2021bottom} &  HRNet-w32    & \xmark & \xmark & 37.9  & \xmark & \xmark &  36.5 \\  %from CID paper
$\heartsuit$ HrHRNet$^\dag$~\cite{cheng2020higherhrnet} & HRNet-w32  & \xmark &  \xmark & 40.0 & \xmark & \xmark & 39.4 \\  %from CID paper
$\bigstar$ YOLACT~\cite{bolya2019yolact} & R-101-FPN            & \xmark & 13.2 & \xmark  & \xmark & 13.5 & \xmark \\ % from LSNet
$\bigstar$ CondInst~\cite{tian2020conditional} & R-50-FPN           & \xmark & 20.3 & \xmark & \xmark & 20.1 & \xmark \\ % from LSNet
$\bigstar$ LSNet-5~\cite{zhang2021location} & DLA-34              & \xmark & 25.0 & \xmark & \xmark & 24.9 & \xmark \\ % from LSNet
$\bigstar$ LOGO-CAP$^\dag$~\cite{xue2022learning} &  HRNet-w32  & \xmark & \xmark & 39.0 & \xmark & \xmark & 38.1 \\  %from LOGO-CAP paper
$\bigstar$ CID$^\dag$~\cite{wang2022contextual} & R-50-FPN         & \xmark & \xmark & 29.2 & \xmark & \xmark & 28.3 \\  %from mmpose
$\bigstar$ CID$^\dag$~\cite{wang2022contextual} & HRNet-w32        & \xmark & \xmark & 44.9 & \xmark & \xmark & 44.0 \\  %from CID paper\midrule
\rowcolor{ourscolor}
$\bigstar$ HQNet (Ours) & R-50 & 30.6  & 31.5  & 40.3 & 29.5 & 31.1 &  40.0 \\ 
\rowcolor{ourscolor}
$\bigstar$ HQNet (Ours) & ViT-L & 36.9 & 39.9 & 46.8  & 35.8 & 38.8 &  45.6  \\ 
\bottomrule
\end{tabular}
}
\label{tab:ochuman}
\end{table}

\textbf{Comparison with multi-task HCP methods.}
Pose2Seg~\cite{zhang2019pose2seg} is a two-stage human pose-based instance segmentation approach. It uses a standalone keypoint detector for pose estimation and employs human skeleton features for top-down instance segmentation guidance. MultiPoseNet~\cite{abdulnabi2015multi} and PersonLab~\cite{papandreou2018personlab} follow bottom-up strategies. CenterNet~\cite{zhou2019objects}, LSNet~\cite{zhang2021location}, and UniHead~\cite{liang2022unifying}\footnote{UniHead trains separate models for different HCP tasks.} are single-stage alternatives.
Our R-50 model achieves superior performance in multi-task HCP, without bells and whistles.

\textbf{Effect of multi-task co-learning.} In Table~\ref{tab:coco}, we also compare with different variants of HQNet for various task composition (\ie D, S, P, C).
We observed that co-learning with multiple human-centric tasks leads to improved overall performance. This enhancement can be attributed to the inter-task synergy that arises from jointly training different HCP tasks. 

\subsection{Results on the OCHuman Dataset}
To verify the performance of HQNet in challenging crowded scenarios, we compare it with recent works on OCHuman dataset~\cite{zhang2019pose2seg}, which is a crowded scene benchmark for human detection, segmentation, and pose estimation in Table~\ref{tab:ochuman}. We show that our model outperforms previous methods under the same ResNet50 backbone network by a large margin. For instance, it outperforms SBL by 9.6 keypoint AP and CondInst by 11.0 segmentation AP on \texttt{test} set. It even achieves superior performance than HrHRNet (40.3 vs 40.0) and LOGO-CAP (40.3 vs 39.0) even with a much smaller backbone (ResNet-50 vs. HRNet-w32). 
With a stronger backbone, \ie ViT-L, our HQNet sets new state-of-the-art results on detection (35.8 AP), segmentation (38.8 AP), and pose estimation (45.6 AP). 

\begin{table*}[h]
\centering
\begin{minipage}[t]{0.28\textwidth}
\centering
\caption{\textbf{Finetuning evaluation on novel face detection tasks.} Face detection results are reported on COCO-UniHuman \texttt{val} dataset.} 
\vspace{-14pt}
\begin{center}
\small
\scalebox{0.7}{
    \begin{tabular}{l|cc}
        \hline
        \multirow{2}{*}{Method}  & \multicolumn{2}{c}{Face detection}  \\ \cline{2-3}
        &  AP     & AR    \\
        \hline
        Faster RCNN~\cite{renNIPS15fasterrcnn} & 43.9 & 71.2 \\
        ZoomNet~\cite{jin2020whole} & 58.2 & 72.8 \\ 
        \rowcolor{ourscolor}
        HQNet (R-50) & \textbf{68.4} & \textbf{83.2} \\
        \hline
    \end{tabular}
    }
\end{center}
\label{tab:detection}
\end{minipage}
\hspace{1.5mm}
\begin{minipage}[t]{0.33\textwidth}
\centering
\caption{\textbf{Unseen-task evaluation on PoseTrack21~\cite{doering2022posetrack21}.}
    `FT' means fine-tuning on PoseTrack21.
    Our models are evaluated without training on MOT.
    } 
\vspace{-14pt}
\begin{center}\small
\scalebox{0.65}{
    \begin{tabular}{l|c|cc}
        \hline
        Method & FT & IDF1 & MOTA\\
        \hline
        TRMOT~\cite{wang2020towards} & \cmark &  57.3 & 47.2  \\
        FairMOT~\cite{zhang2021fairmot} & \cmark & 63.2 & 56.3  \\ 
        \hline
        \rowcolor{ourscolor}
        HQNet (D) & \xmark & 62.4 & 48.6 \\
        \rowcolor{ourscolor}
        HQNet (D+S) & \xmark & 63.3 & 49.5 \\
        \rowcolor{ourscolor}
        HQNet (R-50) & \xmark & 64.6 & 51.1 \\
        \rowcolor{ourscolor}
        HQNet (ViT-L) & \xmark & \textbf{69.1} & \textbf{57.0} \\
        \hline
    \end{tabular}
    }
\end{center}
\label{tab:mot}

\label{tab:sampling_baseline}
\end{minipage}
\hspace{1mm}
\begin{minipage}[t]{0.33\textwidth}
\centering
\caption{\textbf{Robustness to domain shift.}
    All models are evaluated on Human-Art~\cite{ju2023human} \texttt{val} set without training on Human-Art.
    } 
\vspace{-10pt}
\begin{center}
\small
\scalebox{0.64}{
    \begin{tabular}{l|cc}
        \hline
        Method & Det.& Kpt. \\
        \hline
        Faster R-CNN~\cite{renNIPS15fasterrcnn} + HRNet~\cite{sun2019deep} & 12.0 & 22.2 \\
        YOLOX~\cite{ge2021yolox} + ViTPose~\cite{xu2022vitpose} & 14.4 & 28.7  \\
        HigherHRNet~\cite{cheng2020higherhrnet} & -  &  34.6  \\ 
        ED-Pose~\cite{yang2023explicit} &  - & 37.5  \\
        \rowcolor{ourscolor}
        HQNet (Swin-L) & 15.8 & 43.0  \\
        \rowcolor{ourscolor}
        HQNet (ViT-L) & \textbf{18.7} & \textbf{52.2}  \\
        \hline
    \end{tabular}
    }
\end{center}
\label{tab:humanart}
\end{minipage}
\end{table*}

\subsection{Generalize to New HCP Tasks}
\textbf{Finetuning evaluation.} Similar to linear probing in image classification, we freeze our backbone and transformer encoder (from Table~\ref{tab:coco}) and finetune other parts to evaluate the generalization ability of HQNet on a new HCP task, \ie face detection. In Table~\ref{tab:detection}, we compare our approach with Faster R-CNN~\cite{renNIPS15fasterrcnn} and ZoomNet~\cite{jin2020whole}. 
Our HQNet can not only better exploit the inherent multi-level structure of the human body, but also preserve the efficiency of single-stage detection. 
It outperforms Faster R-CNN (68.4 AP vs 43.9 AP) and ZoomNet (68.4 AP vs 58.2 AP) by a large margin.

\textbf{Unseen-task generalization.} 
We evaluate the generalization ability of our approach through an unseen task evaluation, specifically multiple object tracking (MOT) on the PoseTrack21 dataset~\cite{doering2022posetrack21}. Our models are trained solely on the COCO-UniHuman image-based dataset without explicit tuning for MOT. We hypothesize that our learned human query embeddings, which encode instance-specific features of diverse granularity, can serve as strong cues for distinguishing different objects. We utilized DeepSORT~\cite{wojke2017simple} and used the learned Human Query as re-identification features for association. In Table~\ref{tab:mot}, we compare our results with two state-of-the-art single-network MOT methods that were pretrained on COCO and fine-tuned on PoseTrack21. Despite not explicitly being trained for MOT, our HQNet (R-50) achieves highly competitive results (64.6 IDF1 and 51.1 MOTA). This demonstrates the generalization ability of our learned Human Query. HQNet (D) and HQNet (D+S) refers to HQNet trained solely on the detection (D) and segmentation (S) tasks respectively, and we observed that co-training on multiple HCP tasks improved the quality of the query embeddings (64.6 IDF1 vs. 62.4 IDF1). Furthermore, by employing a stronger ViT backbone, our approach achieves state-of-the-art performance.

\subsection{Robustness to Domain Shift}
HumanArt~\cite{ju2023human} contains images from both natural and artificial (\eg cartoon and painting) scenarios, which can be used for evaluating the robustness to domain shift. In Table~\ref{tab:humanart}, we conduct a system-level cross-domain evaluation by directly evaluating all models on Human-Art  \texttt{val} set without any finetuning. We observe that all models, particularly two-stage models, experienced a decline in performance when a domain gap was present. However, our approach maintained competitive performance, showcasing its resilience to the domain gap.

\subsection{More Analysis}

\begin{figure}[t]
    \centering
    \vspace{-4mm}
    \begin{minipage}{0.43\textwidth}
    \centering
    \includegraphics[width=0.92\textwidth]{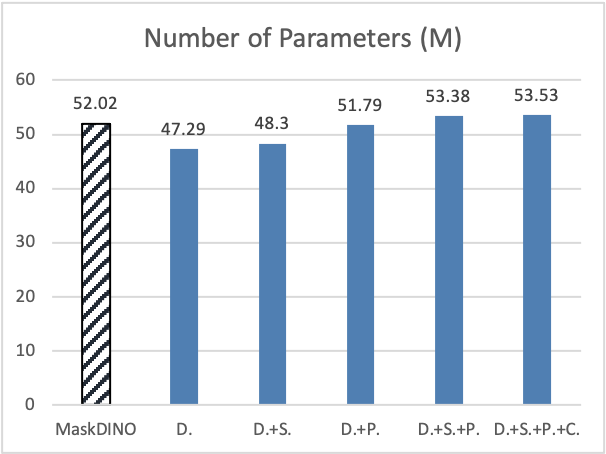}
    \vspace{5mm}
    \caption{Computation cost analysis validates the efficiency of HQNet.} 
  \label{fig:param}
    \end{minipage}%
    \hspace{4mm}
    \begin{minipage}{0.46\textwidth}
\centering
    \includegraphics[width=0.91\textwidth]{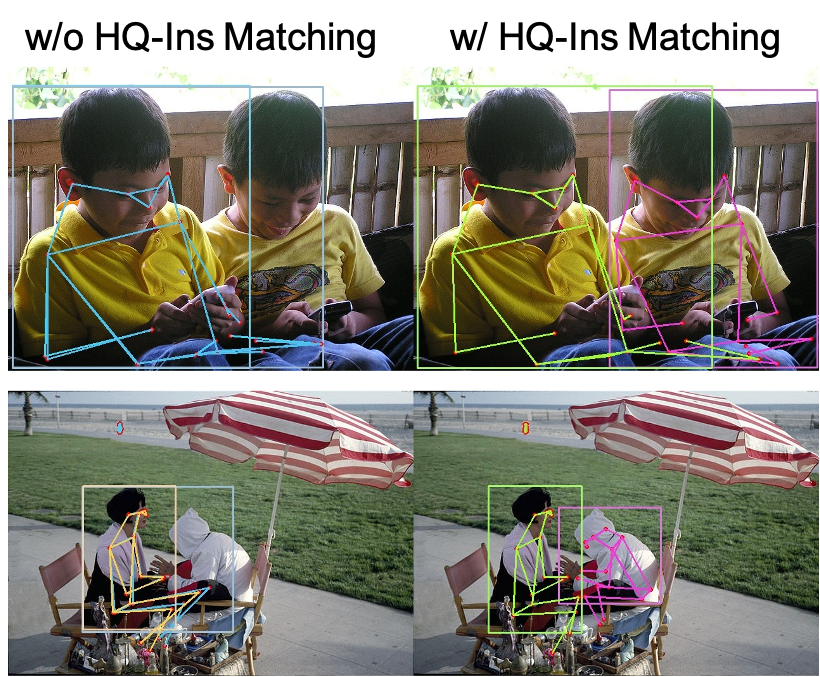}
\caption{Effect of HumanQuery-Instance (HQ-Ins) Matching.
}
\label{fig:hq_ins}
    \end{minipage}
\end{figure}

\textbf{Computation cost analysis.}
In Fig~\ref{fig:param}, we report the number of parameters of our Res50-based HQNet model variants of different task composition. 
In HQNet, multiple tasks share the computation cost of the backbone, transformer encoder and decoder. The overhead of each task-specific head is negligible, showing good scalability of HQNet in terms of increasing the number of tasks. 
It is noteworthy that our model is efficient and its cost is comparable to the task-specific HCP models (\eg MaskDINO~\cite{li2023mask}).

\textbf{Effect of Gender-aided human Model Selection (GaMS)}
In Table~\ref{tab:mesh}, we analyze the effect of Gender-aided human Model Selection (GaMS). We find that incorporating the obtained gender information can assist in selecting proper 3D model of the human body, resulting in more accurate human mesh recovery.

\textbf{Effect of HQ-Ins Matching.} 
``w/o HQ-Ins Matching'' means using detection loss only for bipartite matching~\cite{zhang2022dino}. ``w/ HQ-Ins Matching'' means comprehensively using detection, pose, and segmentation loss for bipartite matching.
As shown in Fig.~\ref{fig:hq_ins}, with detection only matching~\cite{zhang2022dino}, there may be some erroneous cases when one person's pose is matched to another person. HQ-Ins Matching avoids such errors by comprehensively considering multiple tasks as a whole. 
More quantitative evaluation can be found in Supplementary.

%% file: sec/6_conclusion.tex
\section{Conclusion}
In this work, we present a unified solution towards single-stage multi-task human-centric perception, called HQNet. The core idea is to learn a unified query representation that encodes local and global appearance features, coarse and fine-grained localization features for each instance. 
To facilitate model training and evaluation, we introduce a large-scale benchmark, termed COCO-UniHuman benchmark, to unify different representative HCP tasks.
We extensively compare our proposed method with several state-of-the-art task-specific and multi-task approaches, and show the effectiveness of our proposed method.

\noindent\textbf{Acknowledgement.}
This paper is partially supported by the National Key R\&D Program of China No.2022ZD0161000 and the General Research Fund of Hong Kong No.17200622 and 17209324.

%% file: sec/supp.tex
\section{COCO-UniHuman Dataset Statistics}

\begin{figure}[htb]
\centering
    \includegraphics[width=0.6\textwidth]{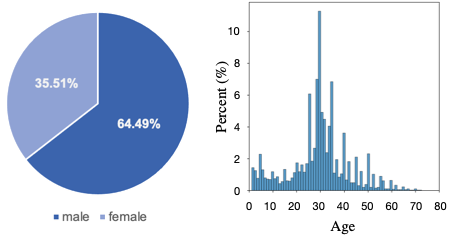}
\caption{Statistics of the COCO-UniHuman benchmark. (a) The gender distribution of COCO-UniHuman is biased towards male. (b) The age distribution ranges from [1, 84] and is biased towards young adults, since images are from public Internet repositories.
}
\label{fig:stat}
\end{figure}

In Fig.~\ref{fig:stat}, we show statistics of our proposed COCO-UniHuman dataset. The plots show the distribution of the gender and the apparent age. We find gender and age biases existed in the widely used COCO dataset. The occurrence of men is significantly higher than women in COCO dataset. More specifically, male to female ratio is about 65:35. In addition, the dataset has an unbalanced age distribution. The apparent age distribution ranges from [1, 84], and it is mainly concentrated between the ages of 25 and 35. 
Analyzing and addressing the gender and age bias in the computer vision system can also be an important topic in the AI community. Future work could also use our benchmark dataset to comprehensively measure and analyze such biases, but it is out of the scope of this paper.

\section{COCO-UniHuman Dataset Annotation}

Obtaining the reliable apparent age is challenging even for human perception. The apparent age will be influenced not only by the real age, but also by other biological and sociological factors of ``aging''. Therefore, there are significant variations on appearance among people of the same age. In this work, we propose the body-based and two-stage annotation strategy to improve the age annotation quality. We also conduct some experiments to show the effectiveness of the proposed age annotation strategy.

\subsection{Body-based vs face-based annotation strategies.} 
\label{sec:body-based}
In this study, we design experiments to compare three different annotation strategies. (1) face-based without face alignment, where the annotation is based on the cropped face image, (2) face-based with face alignment, where face cropping and face alignment pre-processing~\cite{zhang2017quantifying} is applied before annotation, (3) body-based, where the annotation is based on the cropped body image. 
We randomly selected 500 sample person images, and applied the aforementioned 3 strategies to process the data individually, and obtained 3 data sets. 
We also randomly divided 30 well-trained annotators into three groups of 10 annotators each. Each data set was labeled by one group of annotators. 
Each annotator was asked to independently give votes of apparent age for the whole data set. 
As a result, for each body or face image, we have 10 votes. We take the average of the 10 votes as the ground-truth age annotation, and calculate the Age-5 and Age-10 consistency separately. Age-n consistency is defined as:
\begin{equation}
    \frac{1}{K \times N} \sum_{\substack{1 \leq i \leq N \\ 1 \leq j \leq K}}{\mathbb{I}\{|x_{i,j}-x_i^*| \leq n}\} \times 100\%,
\end{equation}
where $N=500$ is the total number of images, and $K=10$ is the number of votes for each image. $x_{i,j}$ is the $j$-th vote for the $i$-th image, while $x_i^*$ means the ground-truth age annotation for the $i$-th image.

\begin{table}[htb]
    \centering
    \caption{Comparisons of age annotation strategies.}
    \setlength{\tabcolsep}{10pt}
    \begin{tabular}{c|c|c}
    \hline
    Annotation Strategy  & Age-5 & Age-10 \\
    \hline
    face w/o alignment & 75.3 & 93.5 \\
    face w/ alignment~\cite{zhang2017quantifying} & 78.2 & 96.5 \\
    body & \textbf{80.9} & \textbf{98.1} \\ 
    \hline
    \end{tabular}
    \label{tab:age_annotation}
\end{table}

From Table~\ref{tab:age_annotation}, we find that the body-based age annotation is better than the face-based age annotation, indicating that the whole-body image contains richer visual cues for age estimation. Interestingly, we also find that face alignment will help improve the age estimation consistency even for human annotators.

\subsection{Two-stage vs one-stage annotation strategies.}
\label{sec:two-stage}
In this study, we design experiments to compare the two-stage and one-stage annotation strategies. For one-stage annotation, we directly annotate the apparent age of the subject. For two-stage annotation, we first annotate the age group and then label the apparent age based on the age group. Table~\ref{tab:age_anno_stage}, shows that two-stage annotation strategy improves the annotation consistency. 

\begin{table}[htb]
    \centering
    \caption{Effect of two-stage age annotation.}
    \setlength{\tabcolsep}{10pt}
    \begin{tabular}{c|c|c}
    \hline
    Annotation Strategy  & Age-5 & Age-10 \\
    \hline
    One-stage age annotation & 80.9 & 98.1 \\ 
    Two-stage age annotation & \textbf{82.1} & \textbf{98.5} \\ 
    \hline
    \end{tabular}
    \label{tab:age_anno_stage}
\end{table}

\section{More Experimental Analysis}

\begin{figure}[htb]
\centering
    \includegraphics[width=0.5\textwidth]{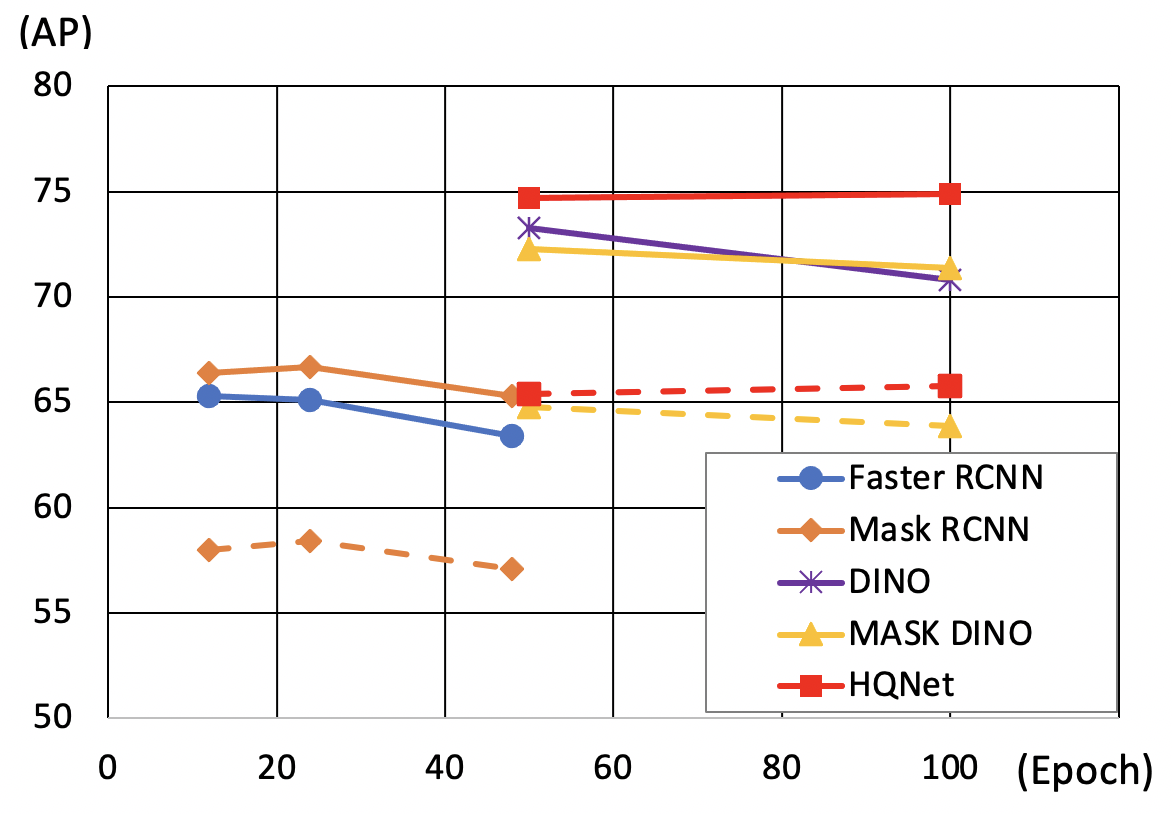}
\caption{Results of human detection (solid lines) and segmentation (dashed lines) with different training schedules on COCO-UniHuman \texttt{val} set. For RCNN-based models, we choose 1x, 2x, and 4x training settings. For DETR-based models, we use 50-epoch and 100-epoch training settings. 
}
\label{fig:overfit}
\end{figure}

\textbf{Multi-task co-learning can mitigate over-fitting.} 
From Fig.~\ref{fig:overfit}, we observe that training task-specific models on ``Person'' category only will easily lead to over-fitting problem, the performance decreases with increasing number of epochs. Specifically, in this experiment, we compare the common 1x, 2x, and 4x training settings for RCNN-based methods (\ie Faster-RCNN, Mask RCNN), and compare 50-epoch and 100-epoch settings for DETR-based methods (\ie DINO, Mask DINO, and our HQNet). The models are trained using MMDetection~\cite{chen2019mmdetection} with suggested hyper-parameters. We report the Average Precision (AP) for both human detection (solid lines) and the human instance segmentation (dashed lines) on COCO-UniHuman \texttt{val} set. Interestingly, we find that our presented multi-task co-learning (HQNet) can mitigate the over-fitting problem, and the performance consistently improves with the increasing training epochs, demonstrating good scalability.

\begin{table}[tb]
\caption{Comparison of general 80 class models and person-specific models on the COCO-UniHuman \texttt{val} set. We report AP for the ``Person'' category without \textit{Small} category person. ``R'' is ResNet~\cite{he2016deep}, and ``FPN'' is feature pyramid network~\cite{lin2017feature}.
The asterisk~* denotes models trained to handle general 80 classes. }
\centering
\small
\scalebox{0.83}{
\setlength{\tabcolsep}{12pt}
\begin{tabular}{l c c c c c c c}
\toprule
\multirow{2}{*}{Model} & \multirow{2}{*}{Backbone} & \multicolumn{3}{c}{Det.} & \multicolumn{3}{c}{Seg.} \\
\cmidrule(r){3-5}
\cmidrule(r){6-8}
& & $\text{AP}$ & $\text{AP}^{M}$ & $\text{AP}^{L}$ &  $\text{AP}$  & $\text{AP}^{M}$ & $\text{AP}^{L}$ \\
\toprule
Faster R-CNN$^*$  & R-50 & 63.0 & 59.8 & 68.1 & \xmark & \xmark & \xmark  \\
Faster R-CNN  & R-50 & 65.3 & 61.5 & 71.2  & \xmark & \xmark & \xmark  \\
Mask R-CNN$^*$ & R-50-FPN & 64.1 & 60.5 & 69.6 & 56.3 & 50.1  & 63.9   \\
Mask R-CNN & R-50-FPN &  66.7 &  62.3 & 73.1 & 58.4 & 51.8 & 66.2  \\ 
\bottomrule
\end{tabular}
}
\label{tab:general_vs_person}
\end{table}

\subsection{General class models vs person-specific models}
In Table~\ref{tab:general_vs_person}, we compare general 80-class models and person-specific models on COCO-UniHuman \texttt{val} set. We find that person-specific models achieve slightly better performance than the general 80-class models for human analysis.  
The asterisk * denotes models trained to handle general 80 classes. All models are evaluated on ``Person'' category without \emph{Small} person. As shown in previous section, training on ``Person'' category only may lead to over-fitting problem. In the experiments, we report the best result for these baseline models. Specifically, Faster R-CNN is trained for 1x, Mask R-CNN for 2x. More details can be found in the section of ``Details about Baseline Models'' below.

\subsection{Effect of HumanQuery-Instance Matching} 
\label{sec:exp_hq_ins}

In Table~\ref{tab:matching}, we quantitatively analyze the effect of HumanQuery-Instance (HQ-Ins) Matching on COCO-UniHuman \texttt{val} set using the ResNet-50 backbone. Note that we use the standard 100-epoch training setting in the experiment. 
We report AP for `Det', `Seg', `Kpt', `Gener', and `Age', which represent detection, keypoint estimation, instance segmentation, gender and age estimation respectively. We show the effectiveness of our proposed HumanQuery-Instance Matching in making the optimization of multi-task HCP learning more consistent and achieving better balance of multiple human-centric analysis tasks.

\begin{table*}[h]
\caption{Effect of HumanQuery-Instance Matching. Experiments are conducted on COCO-UniHuman \texttt{val} set using the ResNet-50 backbone with 100-epoch training setting. We report AP for the ``Person'' category without \textit{Small} category person.}
\centering
\small
\scalebox{0.9}{
\begin{tabular}{c c c| c c c c c c c c c c c c c c c}
\toprule
\multicolumn{3}{c|}{Matching} & \multicolumn{3}{c}{Det.} &  \multicolumn{3}{c}{Seg.} & \multicolumn{3}{c}{Pose (Kpt.)}  & \multicolumn{3}{c}{Cls. (Gender)}  & \multicolumn{3}{c}{Cls. (Age)}\\
\cmidrule(r){4-6}
\cmidrule(r){7-9}
\cmidrule(r){10-12}
\cmidrule(r){13-15}
\cmidrule(r){16-18}
 Box&  Pose & Mask & $\text{AP}$ & $\text{AP}^{M}$ & $\text{AP}^{L}$ &  $\text{AP}$  & $\text{AP}^{M}$ & $\text{AP}^{L}$ &  $\text{AP}$  & $\text{AP}^{M}$ & $\text{AP}^{L}$ &  $\text{AP}$  & $\text{AP}^{M}$ & $\text{AP}^{L}$  &  $\text{AP}$  & $\text{AP}^{M}$ & $\text{AP}^{L}$\\
\toprule
\checkmark & & & 76.2 & 71.4 & 82.4 & 66.1 & 59.1 & 74.2 & 66.8 & 61.0 & 75.0 & 52.1 & 37.3 & 60.7 & 54.0 & 41.2 & 62.0 \\ 
\checkmark & \checkmark & &74.4 & 70.2 & 80.1 & 65.5 & 58.5 & 73.2 & 69.0 & 63.9 & 76.4 & 54.4 & 39.9 & 62.7 & 55.9 & 42.0 & 63.9 \\
\checkmark & \checkmark & \checkmark & 74.9 & 70.4 & 80.7 & 65.8 & 58.7 & 73.9 & 69.3 & 63.8 & 77.3 & 53.8 & 39.7 & 61.2 & 56.0 & 42.5 & 63.3 \\
\bottomrule
\end{tabular}
}
\label{tab:matching}
\end{table*}

\subsection{Qualitative Results}

\begin{figure*}[tb]
\centering
    \includegraphics[width=0.99\textwidth]{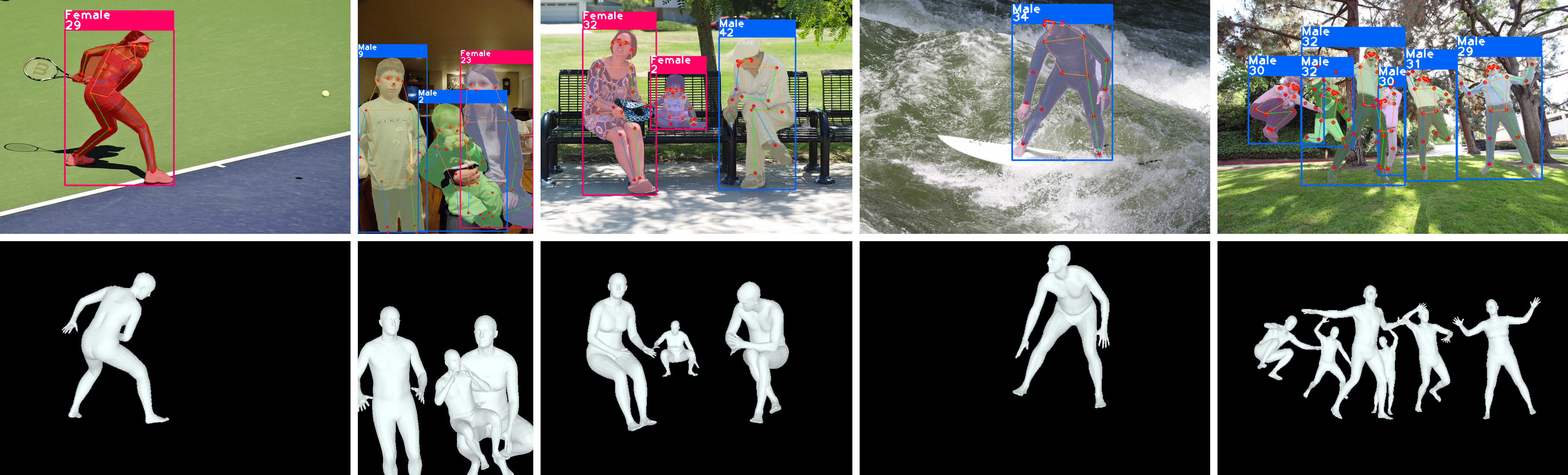}
    \caption{Qualitative results on COCO-UniHuman \texttt{val} dataset. Our HQNet achieves accurate human detection, human pose estimation, human instance segmentation, human attribute recognition and human mesh estimation simultaneously.
}
\label{fig:vis_coco}
\end{figure*}

\begin{figure*}[tb]
\centering
    \includegraphics[width=0.9\textwidth]{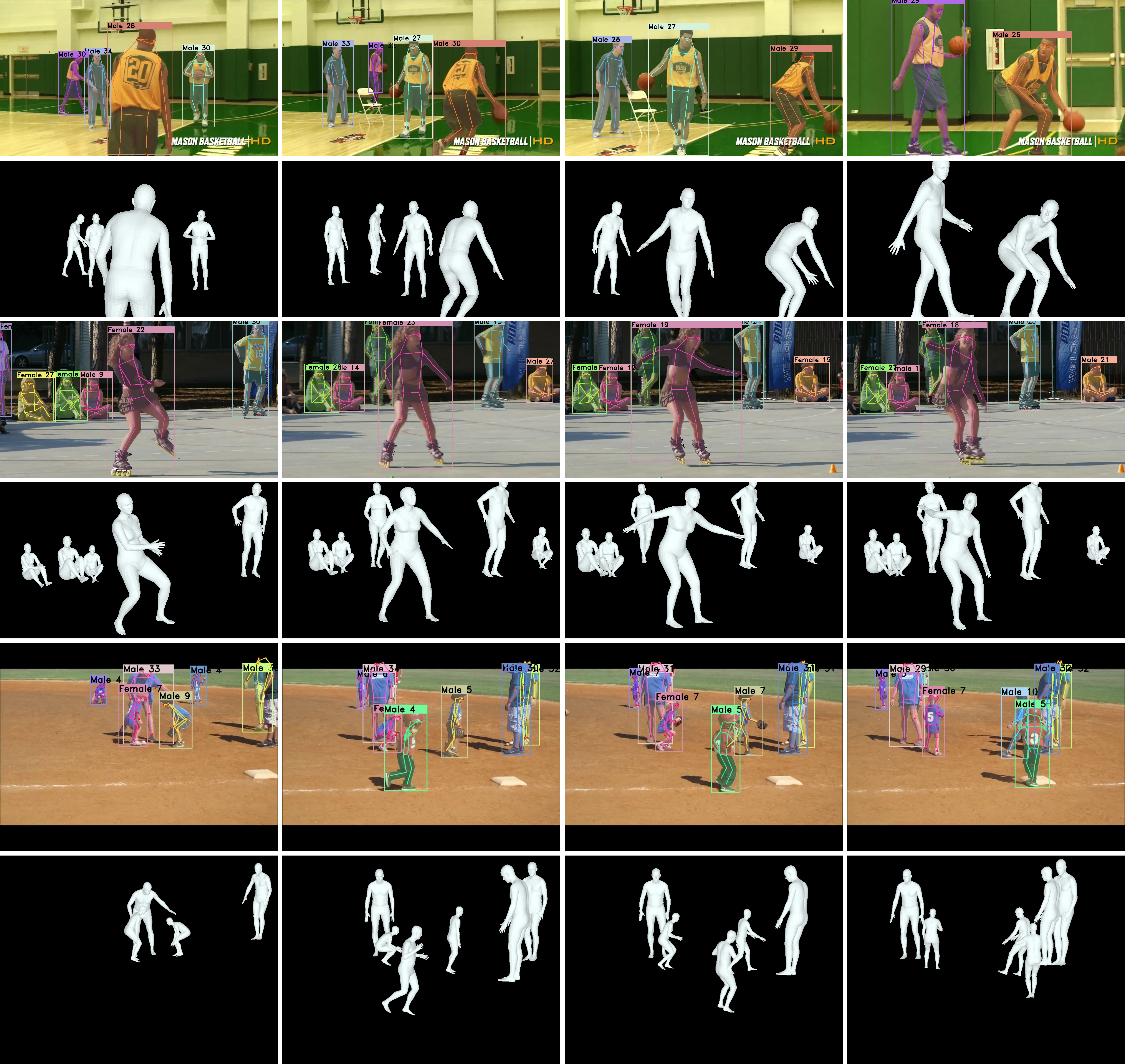}
    \caption{Qualitative results on PoseTrack21 \texttt{val} video dataset. Our HQNet makes a comprehensive human analysis system that can achieve multiple functions: multiple object tracking with human pose estimation, human instance segmentation, and human attribute recognition. Our HQNet is only trained on the COCO-UniHuman image-based dataset without finetuning on the PoseTrack21 video-based dataset.
}
\label{fig:vis_posetrack}
\end{figure*}

In Fig.~\ref{fig:vis_coco} and Fig.~\ref{fig:vis_posetrack}, we show some qualitative results of HQNet with ResNet-50 backbone. 
In Fig.~\ref{fig:vis_coco}, we show some qualitative results on COCO-UniHuman \texttt{val} dataset for human detection, human pose estimation, human instance segmentation and human attribute recognition, and human mesh estimation. The model can recognize the gender and age of different people 
In Fig.~\ref{fig:vis_posetrack}, we visualize the results of human detection and tracking (same color for same id), human pose estimation, human instance segmentation, gender estimation, age estimation and mesh estimation. 
As introduced in the section of ``Unseen-task generalization'' in the main paper, we directly apply our HQNet on multiple object tracking on the challenging PoseTrack21~\cite{doering2022posetrack21} dataset, where our models are trained only on the COCO-UniHuman image-based dataset without explicitly tuned for multi-object tracking (MOT) on video-based dataset like PoseTrack21. Our learned Human Query, which encodes both spatial and visual cues, can serve as good embedding features to distinguish different human instances. Therefore, our human tracking is robust to heavy occlusion, and the id can recover from occlusions. Our HQNet makes a comprehensive all-in-one human analysis system that can achieve multiple functions: multiple object tracking with human pose estimation, human instance segmentation, human attribute recognition and human mesh estimation. 

\subsection{Attention Visualization} 

\begin{figure*}[tb]
\centering
    \includegraphics[width=0.9\textwidth]{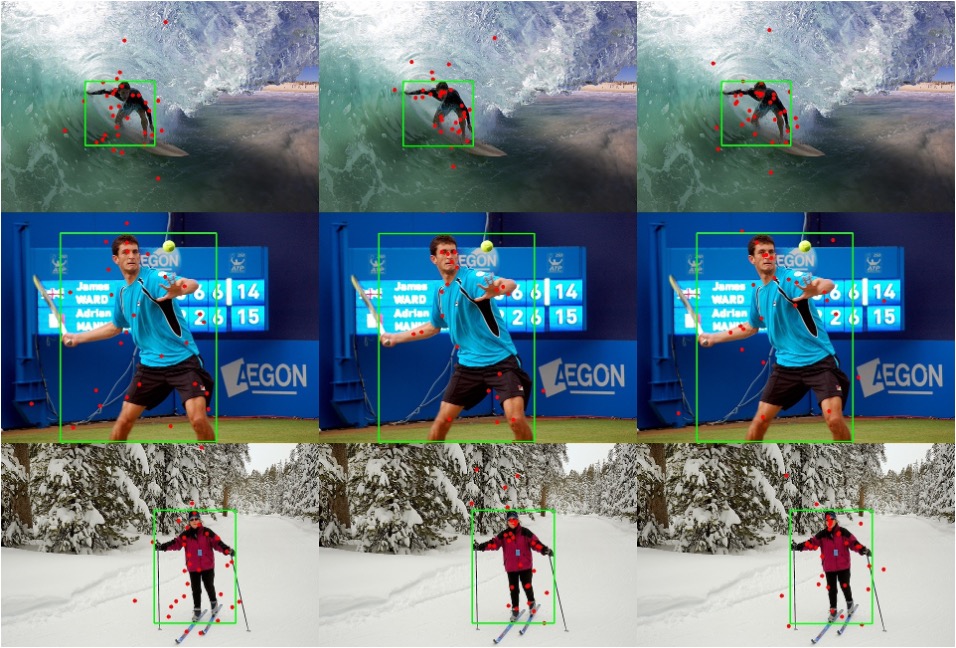}
\caption{Visualization of deformable attention sampling points. The results are from models trained for different HCP tasks. Left: detection and segmentation. Middle: detection and pose. Right: detection, pose, segmentation and attribute. 
}
\label{fig:vis_reference_points}
\end{figure*}

In Fig.~\ref{fig:vis_reference_points}, we visualize the sampling locations of deformable attention for different HCP models. We show the results of the last decoder layer in HQNet-ResNet50. Each sampling point is marked as a red-filled circle. The left results are from the model trained for detection and segmentation ($M_{D+S}$). The middle ones are from the model trained for detection and pose ($M_{D+P}$). And the right ones are from the model trained for detection, segmentation, pose and attribute ($M_{D+P+S+C}$). With the segmentation task, we notice that some of the sampling points of $M_{D+S}$ are distributed near the boundary of the human body, and some are distributed in the background to capture more context information. The sampling points of $M_{D+P}$ have higher probability to distribute inside the human body and some of the points are located closer to the defined human body keypoints, especially the face, arms, and legs. $M_{D+P+S+C}$ combines the characteristics of $M_{D+S}$ and $M_{D+P}$.

\subsection{Failure Case Analysis}

\begin{figure*}[t]
\centering
    \includegraphics[width=0.99\textwidth]{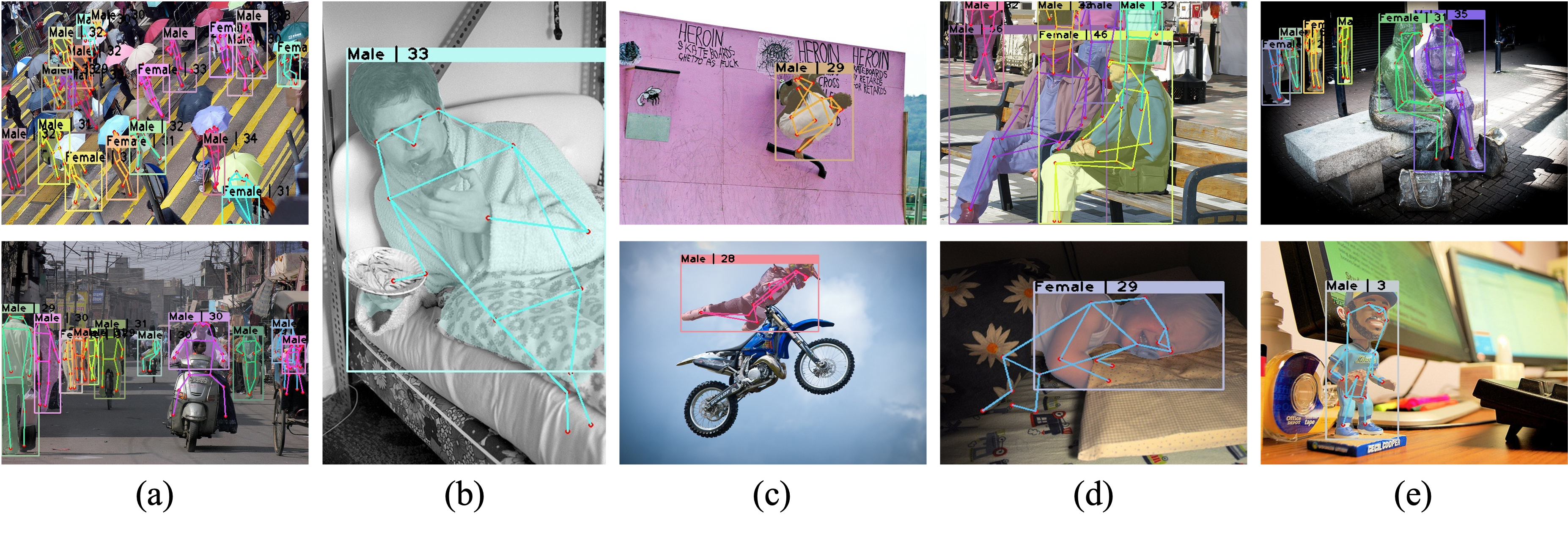}
    \caption{
    Common failure cases: (a) missing detection in crowded scenes, (b) false pose detection in occluded scenes, (c) rare pose or appearance, (d) inaccurate or biased age estimation, (e) false positives on statues or toys.
}
\label{fig:failure}
\end{figure*}

We have analyzed the main cases where our approach fails in the COCO-UniHuman \texttt{val}
set. Fig.~\ref{fig:failure} shows an overview of common failure cases. In highly crowded and occluded scenes, where people are overlapping, the method tends to miss some targets. Occlusion can also lead to pose estimation errors. There will be high keypoint localization errors on non-typical poses (\eg upside-down cases). Due to age/gender bias in the dataset, the method may have some erroneous predictions on human attributes. Statues and toys also frequently lead to false positive errors. 
Most of these issues could be mitigated by adding related data for model training. For example, negative examples could help the network distinguish between humans and other humanoid figures. Adding occluded keypoint annotations could help predict body parts more accurately in occluded scenes.

\section{More Implementation Details}

\subsection{Loss Functions}
\label{sec:appendix_loss_function}

In this work, we jointly train multiple human-centric perception (HCP) tasks, including human detection, human instance segmentation, human pose estimation human attribute (gender and age) recognition and human mesh estimation. 

For human detection, we follow DINO~\cite{zhang2022dino} to apply focal loss~\cite{lin2017focal} for classification $L^{focal}_{cls}$ and detection loss (L1 regression loss $L^{reg}_{det}$ and GIOU loss~\cite{rezatofighi2019generalized} $L^{giou}_{det}$).
For human pose estimation, we follow PETR~\cite{shi2022end} to use focal loss for classifying valid and invalid human instances $L^{focal}_{kpt}$, 
L1 keypoint regression loss $L^{reg}_{kpt}$, OKS loss $L^{oks}_{kpt}$, and auxiliary heatmap loss $L^{hm}_{kpt}$. 
For segmentation loss, we use binary cross-entropy loss $L^{bce}_{seg}$ and dice loss $L^{dice}_{seg}$.
For attribute recognition, we have binary cross-entropy loss for gender estimation $L^{bce}_{gender}$ and mean-variance loss~\cite{pan2018mean} for age estimation $L^{mean}_{age}$ and $L^{var}_{age}$.
For mesh estimation, we use L1 regression loss for pose estimation $L^{reg}_{pose}$, shape estimation $L^{reg}_{shape}$ and the 3D joints regressed from the body model $L^{reg}_{3d}$.

Formally, the overall loss function can be formulated as a linear combination of these sub-task loss functions:
\begin{align*}
    L & = \lambda^{focal}_{cls} L^{focal}_{cls} + \lambda^{reg}_{det} L^{reg}_{det} + \lambda^{giou}_{det} L^{giou}_{det}  \\ 
    & + \lambda^{focal}_{2d} L^{focal}_{2d} + \lambda^{reg}_{2d} L^{reg}_{2d} + \lambda^{oks}_{2d} L^{oks}_{2d} + \lambda^{hm}_{2d} L^{hm}_{2d} \\
    & + \lambda^{reg}_{pose} L^{reg}_{pose} + \lambda^{reg}_{shape} L^{reg}_{shape} + \lambda^{reg}_{3d} L^{reg}_{3d} \\
    & + \lambda^{bce}_{seg} L^{bce}_{seg} + \lambda^{dice}_{seg} L^{dice}_{seg} \\
    & + \lambda^{bce}_{gender} L^{bce}_{gender} + \lambda^{mean}_{age} L^{mean}_{age} + \lambda^{var}_{age} L^{var}_{age},
,
\end{align*}
where $\lambda$s are corresponding loss weights. Detailed settings for the loss weights can be found in Table~\ref{tab:loss_weight}.

\begin{table}[htb]
    \centering
    \caption{Loss weights for training our models.}
    \scalebox{0.99}{
    \setlength{\tabcolsep}{10pt}
    \begin{tabular}{c|c|c}
    \toprule
    \multirow{4}{*}{Detection}       & $\lambda^{focal}_{cls}$ & 1.0 \\ 
    \cmidrule(r){2-3}
                                     & $\lambda^{reg}_{det}$ & 5.0 \\ 
    \cmidrule(r){2-3}
                                     & $\lambda^{giou}_{det}$ & 2.0 \\
    \cmidrule(r){1-3}
    \multirow{8}{*}{Pose}        & $\lambda^{focal}_{kpt}$ & 1.0 \\
    \cmidrule(r){2-3} 
                                     & $\lambda^{reg}_{kpt}$ & 50.0 \\
    \cmidrule(r){2-3} 
                                     & $\lambda^{oks}_{kpt}$ & 1.5 \\
    \cmidrule(r){2-3} 
                                     & $\lambda^{hm}_{kpt}$ & 4.0 \\
    \cmidrule(r){2-3} 
                                     & $\lambda^{reg}_{pose}$ & 5.0 \\ 
    \cmidrule(r){2-3} 
                                     & $\lambda^{reg}_{shape}$ & 10.0 \\ 
    \cmidrule(r){2-3} 
                                     & $\lambda^{reg}_{3d}$ & 10.0 \\ 
    \cmidrule(r){1-3}
    \multirow{3}{*}{Segmentation}    & $\lambda^{bce}_{seg}$ & 8.0 \\ 
    \cmidrule(r){2-3} 
                                     & $\lambda^{dice}_{seg}$ & 5.0 \\
    \cmidrule(r){1-3}
    \multirow{4}{*}{Attribute}       & $\lambda^{bce}_{gender}$ & 1.0 \\ 
    \cmidrule(r){2-3} 
                                     & $\lambda^{mean}_{age}$ & 0.002 \\ 
    \cmidrule(r){2-3} 
                                     & $\lambda^{var}_{age}$ & 0.01 \\ 
    \bottomrule
    \end{tabular}
    }
    \label{tab:loss_weight}
    \vspace{-3mm}
\end{table}

\subsection{Details about Training}
We follow the setting of DINO~\cite{zhang2022dino} to augment the input image by random crop, random flip, and random resize. Specifically, we randomly resize the input image to have its shorter side between 480 and 800 pixels and its longer side less or equal to 1333.
The models are trained with AdamW optimizer~\cite{kingma2014adam} with base learning rate of $1 \times 10^{-4}$, momentum of $0.9$ and weight decay of $1 \times 10^{-4}$. 
For all experiments, the models are trained for 100 epochs with a total batch size of 16 and the initial learning rate is decayed at 80th epoch by a factor of 0.1. We use 16 Tesla V100 GPUs for model training.

In the experiments, we report results of three different backbones: the ResNet-50 backbone is pre-trained on ImageNet-1K dataset, Swin-L backbone is pre-trained on ImageNet-22K dataset, and ViT-L backbone whose pre-trained weights are from~\cite{tang2023humanbench}. 
Unlike DINO and Mask DINO which also pre-train models on Objects365~\cite{zhou2019objects}, we only use COCO-UniHuman data for training without Objects365 dataset. For all backbones, we use 4 scales of feature maps feeding to the encoder and an additional high-resolution feature map for mask prediction. In contrast, DINO and MaskDINO use 5 scales for Swin-L models.
Following the common practice in DETR-like models~\cite{li2022dn,zhang2022dino}, we use a 6-layer Transformer encoder and a 6-layer Transformer decoder and 256 as the hidden feature dimension. We use 300 queries and 100 CDN pairs for training. 
Following ~\cite{zhu2020deformable}, we use independent auxiliary
heads to refine the multi-task predictions at each decoder layer. 

\subsection{Details about Inference}
During inference, the input image is resized to have its shorter side being 800 and longer side at most 1333.
All reported numbers are obtained without model ensemble or test-time augmentations (\eg flip test and multi-scale test).

\section{Details about Baseline Models}
\label{sec:baselines}
\textbf{Details about human detection baselines.} For human detection, we compare several baseline approaches, \ie Faster-RCNN~\cite{renNIPS15fasterrcnn}, IterDETR~\cite{zheng2022progressive} and DINO~\cite{zhang2022dino}. Note that Faster-RCNN and DINO are originally trained to handle general 80 classes (marked with * in Table~\ref{tab:general_vs_person}). For fair comparisons, we use MMDetection~\cite{chen2019mmdetection} to re-train and evaluate them on ``Person'' category using the default experimental setting. Note that MMDetection re-implementation can be a little bit better than the original implementation.

\textbf{Details about human pose estimation baselines.} For human pose estimation, we compare with several representative top-down methods (SBL~\cite{xiao2018simple}, HRNet~\cite{sun2019deep}, Swin~\cite{liu2021swin}, ViTPose~\cite{xu2022vitpose} and PRTR~\cite{li2021pose}), bottom-up approaches (HrHRNet~\cite{cheng2020higherhrnet}, DEKR~\cite{geng2021bottom}, and SWAHR~\cite{luo2021rethinking}) and single-stage approaches (FCPose~\cite{mao2021fcpose}, InsPose~\cite{shi2021inspose}, PETR~\cite{shi2022end} and CID~\cite{wang2022contextual}). Note that the results of Swin (Swin-L), ViTPose (ViT-L) and CID (R-50-FPN) are from MMPose~\cite{mmpose2020}, and other results are from their original papers. 
\emph{Top-down methods} generally yield superior performance, but often rely on a separate human detector, incurring redundant computational costs. 
Specifically, SBL, HRNet, Swin and ViTPose use the same person detector provided by~\cite{xiao2018simple}, which is a strong Faster-RCNN~\cite{renNIPS15fasterrcnn} based detector with detection AP $56.4$ for the ``Person'' category on the COCO'2017 \texttt{val} set. PRTR applies a DETR-based person detector for human detection, which achieves $50.2$ AP for the whole ``Person'' category on the COCO'2017 \texttt{val} set. While PRTR introduces an end-to-end variant (E2E-PRTR) optimizing detection and pose jointly, it lags behind separately trained top-down approaches. 
For pose estimation, the input resolution for SBL, HRNet, and Swin is set as $256 \times 192$, while the input resolution for PRTR is $384 \times 288$.
\emph{Bottom-up methods} learn instance-agnostic keypoints and then cluster them into corresponding individuals. HrHRNet, DEKR, and SWAHR adopt the strong HRNet-w32~\cite{sun2019deep} backbone network with an input resolution of $512 \times 512$. 
\emph{Single-stage approaches} directly predict human body keypoints in a single stage. FCPose, InsPose and PETR adopt R-50~\cite{he2016deep} backbone network. The input images are resized to have their shorter sides being 800 and their longer sides less or equal to 1333.
For CID, we report both the results of R-50-FPN and HRNet-w32 backbones. The input resolution of CID is $512 \times 512$. 

\textbf{Details about human instance segmentation baselines.}
For human instance segmentation, we contrast HQNet with state-of-the-art general and human-specific instance segmentation methods. Mask R-CNN~\cite{he2017mask} is an end-to-end top-down approach that optimizes object detection and instance segmentation jointly. 
Given our one-stage pipeline, we also compare against one-stage methods
, including PolarMask~\cite{xie2020polarmask}, MEInst~\cite{zhang2020mask}, YOLACT~\cite{bolya2019yolact}, and CondInst~\cite{tian2020conditional}. 
Results of PolarMask, MEInst, YOLACT, CondInst are from~\cite{zhang2021location}, which are obtained by re-training and evaluating the models on COCO ``Person'' category only. 
PolarMask encodes the instance mask with coordinates, while MEInst encodes the mask into a compact representation vector. YOLACT and CondInst use a series of global prototypes and linear coefficients to represent instance masks. Instead we learn instance-aware Human Query to decouple each human instance. 

\textbf{Details about gender and age estimation baselines.}
Multi-person gender and age estimation remains under-explored in the literature. We establish baselines using StrongBL~\cite{jia2020rethinking} and Mask R-CNN~\cite{he2017mask}. 
StrongBL is a top-down approach which requires an off-the-shelf human detector. For the detection part, we use a pre-trained Mask RCNN to produce human detection results. And for attribute part, we follow official settings to retrain StrongBL on the COCO-UniHuman dataset. The gender and age models use ResNet50 as the backbone with input resolution $256 \times 192$. 
Mask R-CNN is an end-to-end top-down approach, modified with gender or age branches and retrained using MMDetection with default training settings. 

\textbf{Details about mesh estimation baseline.}
For human mesh estimation, we compare HQNet with state-of-the-art one-stage monocular method ROMP~\cite{sun2021monocular}, and two-stage method HMR~\cite{kanazawa2018end} and HMR+~\cite{pang2022benchmarking}. 
Following official settings, we train these models on COCO-UniHuman with ResNet50 backbone. As two-stage models require human bbox as the input, in the experiment, we use GT bbox for comparisons.

\section{Discussion about Unifying HCP Tasks}

\subsection{General network architecture design} 
There are some attempts to design general network architecture for unifying human-centric perception tasks.
Some works propose to design network backbones for HCP tasks. Both CNN-based (\eg HRNet~\cite{wang2020deep}) and Transformer-based backbone networks (\eg TCFormer~\cite{zeng2022not} are proposed for general human-centric visual tasks. Other works focus on designing network heads to unify different HCP tasks. For example, UniHead~\cite{liang2022unifying} designs a novel perception head with unified keypoint representations that can be used in different HCP tasks. Point-Set Anchors~\cite{wei2020point} designs different point-set anchors to provide task-specific initialization for different HCP tasks. 
Unlike these methods, which employ separate task-specific models for different HCP tasks, we consolidate diverse HCP tasks within a single network.

\subsection{Pre-training on HCP tasks} 
There are also works~\cite{hong2022versatile,chen2023beyond,tang2023humanbench} on pre-training on diverse human-centric tasks with large-scale data. 
HCMoCo~\cite{hong2022versatile} introduces a versatile multi-modal (RGB-D) pre-training framework for single-person pose estimation and segmentation. SOLIDER~\cite{chen2023beyond} presents a self-supervised learning framework to learn a general human representation with more semantic information. HumanBench~\cite{tang2023humanbench} builds a large-scale human-centric pre-training dataset and introduces the projector-assisted pre-training method with hierarchical weight sharing. More recently, UniHCP~\cite{ci2023unihcp} presents a unified vision transformer model to perform multitask pre-training at scale.  It employs task-specific queries for attending to relevant features, but tackles one task at a time. Unlike ours,  our approach simultaneously solves multiple HCP tasks in a single forward pass. 
Our proposed method is different from this pre-training based approach. First, these methods mainly focus on the pre-training stage, and require fine-tuning for the optimal performance on specific down-stream tasks. 
Second, these approaches require large-scale joint training on multiple human-centric perception datasets. This makes it unfair to directly compare with models that train on one specific dataset. In addition, large-scale model training is extremely costly. For example, training of UniHCP requires more than 10,000 GPU hours. 
Third, these methods are designed for single-person human analysis (or top-down human analysis). In comparison, our approach solves multiple HCP tasks in a single-stage multi-task manner.

\subsection{Co-learning on HCP tasks} 
Many works have investigated the correlations between pairs of HCP tasks~\cite{zhang2014panda,tian2015pedestrian,lin2022fp,nie2018mutual,nie2018human}. For example,~\cite{tian2015pedestrian} explore to integrate fine-grained person attribute learning into the pipeline of pedestrian detection.
Mask-RCNN~\cite{he2017mask} extends Faster-RCNN by adding extra keypoint localization or segmentation branch to handle pose estimation and instance segmentation respectively.
Pose2Seg~\cite{zhang2019pose2seg} presents a top-down approach for pose-based human instance segmentation. It uses previously generated poses as input instead of the region proposals to extract features for better alignment and performs the down-stream instance segmentation task. PersonLab~\cite{papandreou2018personlab} adopts a bottom-up scheme and solve pose estimation and instance segmentation by applying a greedy decoding process for human grouping.
We propose a single-stage model that learns a general unified representation to handle all representative human-centric perception tasks simultaneously.

\section{Discussion about Human Attribute Recognition}
Visual recognition of human attributes is an important research topic in computer vision. Among all the human attributes, gender and age are arguably the most popular and representative, which is also our main focus.

\subsection{Dataset}
Human attribute recognition datasets can be classified into two categories, \ie facial attribute recognition datasets and pedestrian attribute recognition datasets. 
Most existing attribute recognition datasets only provide center cropped face (facial attribute recognition) or body (for pedestrian attribute recognition) images, making it not suitable for developing and evaluating multi-person attribute recognition algorithms. In comparison, our proposed COCO-UniHuman preserves the original high resolution image and densely annotates attributes for each human instances. One exception is WIDER-Attr~\cite{li2016human}, which also provides the original images. However, the number of images is relatively small. We hope our dataset can serve as a good alternative benchmark dataset for multi-person human attribute recognition. 

Age estimation datasets can be categorized into three groups, \ie age group classification, real age estimation, and apparent age estimation.  
To our best knowledge, public large-scale pedestrian attribute datasets (\eg WIDER-Attr~\cite{li2016human}, PETA~\cite{deng2014pedestrian}, Market1501-Attr~\cite{lin2019improving,zheng2015scalable}, RAP-2.0~\cite{li2019richly} 
and PA-100K~\cite{liu2017hydraplus}) only have coarse age group annotations. Facial attribute datasets may also have fine-grained apparent (\eg APPA-REAL~\cite{agustsson2017apparent}) or real (\eg MegaAge~\cite{zhang2017quantifying}) age annotations. Apparent age estimation focuses on how old a subject ``looks like'', instead of how old a subject ``really is''. It is considered to be a more practical setting for visual analysis. Our proposed dataset is the first large scale in-the-wild dataset for body-based apparent age estimation. Body-based apparent age estimation is promising especially when the facial image is not captured clear enough (\eg captured in a distance). However, body-based apparent age estimation is under-explored in literature due to lack of dataset. We hope our presented COCO-UniHuman dataset can promote related research.

\subsection{Method}
Human attribute recognition focuses on assigning a set of semantic attributes (\eg gender and age) to each human instance. Typical approaches include global image based~\cite{abdulnabi2015multi,jia2020rethinking}, local parts based~\cite{li2016human}, and visual attention based~\cite{liu2017hydraplus} approaches. Most of them focus on single-human (or top-down) analysis without consider the relationship among different human instances. In comparison, we introduce a single-stage multi-person human attribute (\ie gender and age) recognition approach.

\section{Limitations and Future Work}

While our work focuses on RGB image based HCP tasks, tasks about video data or multi-modality data (\eg IR and Depth) also hold significant potential. We encourage future research to explore more comprehensive multi-task human-centric perception.

%% file: arxiv.bbl
\begin{thebibliography}{100}
\providecommand{\url}[1]{\texttt{#1}}
\providecommand{\urlprefix}{URL }
\providecommand{\doi}[1]{https://doi.org/#1}

\bibitem{abdulnabi2015multi}
Abdulnabi, A.H., Wang, G., Lu, J., Jia, K.: Multi-task cnn model for attribute prediction. IEEE Trans. Multimedia  \textbf{17}(11),  1949--1959 (2015)

\bibitem{agustsson2017apparent}
Agustsson, E., Timofte, R., Escalera, S., Baro, X., Guyon, I., Rothe, R.: Apparent and real age estimation in still images with deep residual regressors on appa-real database. In: IEEE Int. Conf. Auto. Face \& Gesture Recog. pp. 87--94 (2017)

\bibitem{alp2018densepose}
Alp~G{\"u}ler, R., Neverova, N., Kokkinos, I.: Densepose: Dense human pose estimation in the wild. In: IEEE Conf. Comput. Vis. Pattern Recog. (2018)

\bibitem{andriluka2018posetrack}
Andriluka, M., Iqbal, U., Insafutdinov, E., Pishchulin, L., Milan, A., Gall, J., Schiele, B.: Posetrack: A benchmark for human pose estimation and tracking. In: IEEE Conf. Comput. Vis. Pattern Recog. (2018)

\bibitem{andriluka20142d}
Andriluka, M., Pishchulin, L., Gehler, P., Schiele, B.: 2d human pose estimation: New benchmark and state of the art analysis. In: IEEE Conf. Comput. Vis. Pattern Recog. (2014)

\bibitem{bolya2019yolact}
Bolya, D., Zhou, C., Xiao, F., Lee, Y.J.: Yolact: Real-time instance segmentation. In: Int. Conf. Comput. Vis. pp. 9157--9166 (2019)

\bibitem{cao2017realtime}
Cao, Z., Simon, T., Wei, S.E., Sheikh, Y.: Realtime multi-person 2d pose estimation using part affinity fields. In: IEEE Conf. Comput. Vis. Pattern Recog. (2017)

\bibitem{carion2020end}
Carion, N., Massa, F., Synnaeve, G., Usunier, N., Kirillov, A., Zagoruyko, S.: End-to-end object detection with transformers. In: Eur. Conf. Comput. Vis. pp. 213--229 (2020)

\bibitem{chen2019mmdetection}
Chen, K., Wang, J., Pang, J., Cao, Y., Xiong, Y., Li, X., Sun, S., Feng, W., Liu, Z., Xu, J., et~al.: Mmdetection: Open mmlab detection toolbox and benchmark. arXiv preprint arXiv:1906.07155  (2019)

\bibitem{chen2023beyond}
Chen, W., Xu, X., Jia, J., Luo, H., Wang, Y., Wang, F., Jin, R., Sun, X.: Beyond appearance: a semantic controllable self-supervised learning framework for human-centric visual tasks. In: IEEE Conf. Comput. Vis. Pattern Recog. pp. 15050--15061 (2023)

\bibitem{cheng2020higherhrnet}
Cheng, B., Xiao, B., Wang, J., Shi, H., Huang, T.S., Zhang, L.: Higherhrnet: Scale-aware representation learning for bottom-up human pose estimation. In: IEEE Conf. Comput. Vis. Pattern Recog. pp. 5386--5395 (2020)

\bibitem{ci2023unihcp}
Ci, Y., Wang, Y., Chen, M., Tang, S., Bai, L., Zhu, F., Zhao, R., Yu, F., Qi, D., Ouyang, W.: Unihcp: A unified model for human-centric perceptions. In: IEEE Conf. Comput. Vis. Pattern Recog. pp. 17840--17852 (2023)

\bibitem{mmpose2020}
Contributors, M.: Openmmlab pose estimation toolbox and benchmark. \url{https://github.com/open-mmlab/mmpose} (2020)

\bibitem{deng2014pedestrian}
Deng, Y., Luo, P., Loy, C.C., Tang, X.: Pedestrian attribute recognition at far distance. In: ACM Int. Conf. Multimedia. pp. 789--792 (2014)

\bibitem{doering2022posetrack21}
Doering, A., Chen, D., Zhang, S., Schiele, B., Gall, J.: Posetrack21: A dataset for person search, multi-object tracking and multi-person pose tracking. In: IEEE Conf. Comput. Vis. Pattern Recog. pp. 20963--20972 (2022)

\bibitem{dollar2009pedestrian}
Doll{\'a}r, P., Wojek, C., Schiele, B., Perona, P.: Pedestrian detection: A benchmark. In: IEEE Conf. Comput. Vis. Pattern Recog. pp. 304--311 (2009)

\bibitem{dosovitskiy2020image}
Dosovitskiy, A., Beyer, L., Kolesnikov, A., Weissenborn, D., Zhai, X., Unterthiner, T., Dehghani, M., Minderer, M., Heigold, G., Gelly, S., et~al.: An image is worth 16x16 words: Transformers for image recognition at scale. Int. Conf. Learn. Represent.  (2021)

\bibitem{ge2021yolox}
Ge, Z., Liu, S., Wang, F., Li, Z., Sun, J.: Yolox: Exceeding yolo series in 2021. arXiv preprint arXiv:2107.08430  (2021)

\bibitem{geng2021bottom}
Geng, Z., Sun, K., Xiao, B., Zhang, Z., Wang, J.: Bottom-up human pose estimation via disentangled keypoint regression. In: IEEE Conf. Comput. Vis. Pattern Recog. pp. 14676--14686 (2021)

\bibitem{gong2018instance}
Gong, K., Liang, X., Li, Y., Chen, Y., Yang, M., Lin, L.: Instance-level human parsing via part grouping network. In: Eur. Conf. Comput. Vis. pp. 770--785 (2018)

\bibitem{he2017mask}
He, K., Gkioxari, G., Doll{\'a}r, P., Girshick, R.: Mask r-cnn. In: Int. Conf. Comput. Vis. pp. 2961--2969 (2017)

\bibitem{he2016deep}
He, K., Zhang, X., Ren, S., Sun, J.: Deep residual learning for image recognition. In: IEEE Conf. Comput. Vis. Pattern Recog. (2016)

\bibitem{hong2022versatile}
Hong, F., Pan, L., Cai, Z., Liu, Z.: Versatile multi-modal pre-training for human-centric perception. In: IEEE Conf. Comput. Vis. Pattern Recog. pp. 16156--16166 (2022)

\bibitem{insafutdinov2016deepercut}
Insafutdinov, E., Pishchulin, L., Andres, B., Andriluka, M., Schiele, B.: Deepercut: A deeper, stronger, and faster multi-person pose estimation model. In: Eur. Conf. Comput. Vis. pp. 34--50 (2016)

\bibitem{ionescu2013human3}
Ionescu, C., Papava, D., Olaru, V., Sminchisescu, C.: Human3.6m: Large scale datasets and predictive methods for 3d human sensing in natural environments. IEEE Trans. Pattern Anal. Mach. Intell.  \textbf{36}(7),  1325--1339 (2013)

\bibitem{jia2020rethinking}
Jia, J., Huang, H., Yang, W., Chen, X., Huang, K.: Rethinking of pedestrian attribute recognition: Realistic datasets with efficient method. arXiv preprint arXiv:2005.11909  (2020)

\bibitem{jiang2022posetrans}
Jiang, W., Jin, S., Liu, W., Qian, C., Luo, P., Liu, S.: Posetrans: A simple yet effective pose transformation augmentation for human pose estimation. In: Eur. Conf. Comput. Vis. pp. 643--659 (2022)

\bibitem{jin2019multi}
Jin, S., Liu, W., Ouyang, W., Qian, C.: Multi-person articulated tracking with spatial and temporal embeddings. In: IEEE Conf. Comput. Vis. Pattern Recog. pp. 5664--5673 (2019)

\bibitem{jin2020differentiable}
Jin, S., Liu, W., Xie, E., Wang, W., Qian, C., Ouyang, W., Luo, P.: Differentiable hierarchical graph grouping for multi-person pose estimation. In: Eur. Conf. Comput. Vis. pp. 718--734 (2020)

\bibitem{jin2020whole}
Jin, S., Xu, L., Xu, J., Wang, C., Liu, W., Qian, C., Ouyang, W., Luo, P.: Whole-body human pose estimation in the wild. In: Eur. Conf. Comput. Vis. (2020)

\bibitem{jin2024unifs}
Jin, S., Yao, R., Xu, L., Liu, W., Qian, C., Wu, J., Luo, P.: Unifs: Universal few-shot instance perception with point representations. In: Eur. Conf. Comput. Vis. (2024)

\bibitem{joo2021exemplar}
Joo, H., Neverova, N., Vedaldi, A.: Exemplar fine-tuning for 3d human model fitting towards in-the-wild 3d human pose estimation. In: Int. Conf. 3D Vis. pp. 42--52. IEEE (2021)

\bibitem{ju2023human}
Ju, X., Zeng, A., Wang, J., Xu, Q., Zhang, L.: Human-art: A versatile human-centric dataset bridging natural and artificial scenes. In: IEEE Conf. Comput. Vis. Pattern Recog. pp. 618--629 (2023)

\bibitem{kanazawa2018end}
Kanazawa, A., Black, M.J., Jacobs, D.W., Malik, J.: End-to-end recovery of human shape and pose. In: IEEE Conf. Comput. Vis. Pattern Recog. pp. 7122--7131 (2018)

\bibitem{kingma2014adam}
Kingma, D.P., Ba, J.: Adam: A method for stochastic optimization. arXiv preprint arXiv:1412.6980  (2014)

\bibitem{kirillov2017instancecut}
Kirillov, A., Levinkov, E., Andres, B., Savchynskyy, B., Rother, C.: Instancecut: from edges to instances with multicut. In: IEEE Conf. Comput. Vis. Pattern Recog. pp. 5008--5017 (2017)

\bibitem{kong2018recurrent}
Kong, S., Fowlkes, C.C.: Recurrent pixel embedding for instance grouping. In: IEEE Conf. Comput. Vis. Pattern Recog. pp. 9018--9028 (2018)

\bibitem{li2019richly}
Li, D., Zhang, Z., Chen, X., Huang, K.: A richly annotated pedestrian dataset for person retrieval in real surveillance scenarios. IEEE Trans. Image Process.  \textbf{28}(4),  1575--1590 (2019)

\bibitem{li2022dn}
Li, F., Zhang, H., Liu, S., Guo, J., Ni, L.M., Zhang, L.: Dn-detr: Accelerate detr training by introducing query denoising. In: IEEE Conf. Comput. Vis. Pattern Recog. pp. 13619--13627 (2022)

\bibitem{li2023mask}
Li, F., Zhang, H., Xu, H., Liu, S., Zhang, L., Ni, L.M., Shum, H.Y.: Mask dino: Towards a unified transformer-based framework for object detection and segmentation. In: IEEE Conf. Comput. Vis. Pattern Recog. pp. 3041--3050 (2023)

\bibitem{li2017multiple}
Li, J., Zhao, J., Wei, Y., Lang, C., Li, Y., Sim, T., Yan, S., Feng, J.: Multiple-human parsing in the wild. arXiv preprint arXiv:1705.07206  (2017)

\bibitem{li2021pose}
Li, K., Wang, S., Zhang, X., Xu, Y., Xu, W., Tu, Z.: Pose recognition with cascade transformers. In: IEEE Conf. Comput. Vis. Pattern Recog. pp. 1944--1953 (2021)

\bibitem{li2016human}
Li, Y., Huang, C., Loy, C.C., Tang, X.: Human attribute recognition by deep hierarchical contexts. In: Eur. Conf. Comput. Vis. (2016)

\bibitem{liang2022unifying}
Liang, J., Song, G., Leng, B., Liu, Y.: Unifying visual perception by dispersible points learning. In: Eur. Conf. Comput. Vis. pp. 439--456 (2022)

\bibitem{lin2023one}
Lin, J., Zeng, A., Wang, H., Zhang, L., Li, Y.: One-stage 3d whole-body mesh recovery with component aware transformer. In: IEEE Conf. Comput. Vis. Pattern Recog. pp. 21159--21168 (2023)

\bibitem{lin2021end}
Lin, K., Wang, L., Liu, Z.: End-to-end human pose and mesh reconstruction with transformers. In: IEEE Conf. Comput. Vis. Pattern Recog. pp. 1954--1963 (2021)

\bibitem{lin2017feature}
Lin, T.Y., Dollár, P., Girshick, R., He, K., Hariharan, B., Belongie, S.: Feature pyramid networks for object detection (2017)

\bibitem{lin2017focal}
Lin, T.Y., Goyal, P., Girshick, R., He, K., Doll{\'a}r, P.: Focal loss for dense object detection. In: Int. Conf. Comput. Vis. pp. 2980--2988 (2017)

\bibitem{lin2014microsoft}
Lin, T.Y., Maire, M., Belongie, S., Hays, J., Perona, P., Ramanan, D., Doll{\'a}r, P., Zitnick, C.L.: Microsoft coco: Common objects in context. In: Eur. Conf. Comput. Vis. (2014)

\bibitem{lin2022fp}
Lin, Y., Shen, J., Wang, Y., Pantic, M.: Fp-age: Leveraging face parsing attention for facial age estimation in the wild. IEEE Trans. Image Process.  (2022)

\bibitem{lin2019improving}
Lin, Y., Zheng, L., Zheng, Z., Wu, Y., Hu, Z., Yan, C., Yang, Y.: Improving person re-identification by attribute and identity learning. Pattern Recognition  (2019)

\bibitem{liu2022dab}
Liu, S., Li, F., Zhang, H., Yang, X., Qi, X., Su, H., Zhu, J., Zhang, L.: Dab-detr: Dynamic anchor boxes are better queries for detr. Int. Conf. Learn. Represent.  (2022)

\bibitem{liu2017hydraplus}
Liu, X., Zhao, H., Tian, M., Sheng, L., Shao, J., Yan, J., Wang, X.: Hydraplus-net: Attentive deep features for pedestrian analysis. In: Int. Conf. Comput. Vis. pp.~1--9 (2017)

\bibitem{liu2021swin}
Liu, Z., Lin, Y., Cao, Y., Hu, H., Wei, Y., Zhang, Z., Lin, S., Guo, B.: Swin transformer: Hierarchical vision transformer using shifted windows. In: Int. Conf. Comput. Vis. pp. 10012--10022 (2021)

\bibitem{liu2015faceattributes}
Liu, Z., Luo, P., Wang, X., Tang, X.: Deep learning face attributes in the wild. In: Int. Conf. Comput. Vis. (2015)

\bibitem{luo2021rethinking}
Luo, Z., Wang, Z., Huang, Y., Wang, L., Tan, T., Zhou, E.: Rethinking the heatmap regression for bottom-up human pose estimation. In: IEEE Conf. Comput. Vis. Pattern Recog. pp. 13264--13273 (2021)

\bibitem{mao2021fcpose}
Mao, W., Tian, Z., Wang, X., Shen, C.: Fcpose: Fully convolutional multi-person pose estimation with dynamic instance-aware convolutions. In: IEEE Conf. Comput. Vis. Pattern Recog. pp. 9034--9043 (2021)

\bibitem{newell2017associative}
Newell, A., Huang, Z., Deng, J.: Associative embedding: End-to-end learning for joint detection and grouping. In: Adv. Neural Inform. Process. Syst. (2017)

\bibitem{nie2018mutual}
Nie, X., Feng, J., Yan, S.: Mutual learning to adapt for joint human parsing and pose estimation. In: Eur. Conf. Comput. Vis. pp. 502--517 (2018)

\bibitem{nie2019single}
Nie, X., Feng, J., Zhang, J., Yan, S.: Single-stage multi-person pose machines. In: Int. Conf. Comput. Vis. pp. 6951--6960 (2019)

\bibitem{nie2018human}
Nie, X., Feng, J., Zuo, Y., Yan, S.: Human pose estimation with parsing induced learner. In: IEEE Conf. Comput. Vis. Pattern Recog. pp. 2100--2108 (2018)

\bibitem{pan2018mean}
Pan, H., Han, H., Shan, S., Chen, X.: Mean-variance loss for deep age estimation from a face. In: IEEE Conf. Comput. Vis. Pattern Recog. pp. 5285--5294 (2018)

\bibitem{pang2022benchmarking}
Pang, H.E., Cai, Z., Yang, L., Zhang, T., Liu, Z.: Benchmarking and analyzing 3d human pose and shape estimation beyond algorithms. Advances in Neural Information Processing Systems  \textbf{35},  26034--26051 (2022)

\bibitem{papandreou2018personlab}
Papandreou, G., Zhu, T., Chen, L.C., Gidaris, S., Tompson, J., Murphy, K.: Personlab: Person pose estimation and instance segmentation with a bottom-up, part-based, geometric embedding model. In: Eur. Conf. Comput. Vis. pp. 269--286 (2018)

\bibitem{renNIPS15fasterrcnn}
Ren, S., He, K., Girshick, R., Sun, J.: Faster {R-CNN}: Towards real-time object detection with region proposal networks. In: Adv. Neural Inform. Process. Syst. (2015)

\bibitem{rezatofighi2019generalized}
Rezatofighi, H., Tsoi, N., Gwak, J., Sadeghian, A., Reid, I., Savarese, S.: Generalized intersection over union: A metric and a loss for bounding box regression. In: IEEE Conf. Comput. Vis. Pattern Recog. pp. 658--666 (2019)

\bibitem{rothe2015dex}
Rothe, R., Timofte, R., Van~Gool, L.: Dex: Deep expectation of apparent age from a single image. In: Int. Conf. Comput. Vis. Worksh. pp. 10--15 (2015)

\bibitem{shao2018crowdhuman}
Shao, S., Zhao, Z., Li, B., Xiao, T., Yu, G., Zhang, X., Sun, J.: Crowdhuman: A benchmark for detecting human in a crowd. arXiv preprint arXiv:1805.00123  (2018)

\bibitem{shi2022end}
Shi, D., Wei, X., Li, L., Ren, Y., Tan, W.: End-to-end multi-person pose estimation with transformers. In: IEEE Conf. Comput. Vis. Pattern Recog. pp. 11069--11078 (2022)

\bibitem{shi2021inspose}
Shi, D., Wei, X., Yu, X., Tan, W., Ren, Y., Pu, S.: Inspose: instance-aware networks for single-stage multi-person pose estimation. In: ACM Int. Conf. Multimedia. pp. 3079--3087 (2021)

\bibitem{sun2019deep}
Sun, K., Xiao, B., Liu, D., Wang, J.: Deep high-resolution representation learning for human pose estimation. In: IEEE Conf. Comput. Vis. Pattern Recog. pp. 5693--5703 (2019)

\bibitem{sun2021monocular}
Sun, Y., Bao, Q., Liu, W., Fu, Y., Black, M.J., Mei, T.: Monocular, one-stage, regression of multiple 3d people. In: Int. Conf. Comput. Vis. pp. 11179--11188 (2021)

\bibitem{tang2023humanbench}
Tang, S., Chen, C., Xie, Q., Chen, M., Wang, Y., Ci, Y., Bai, L., Zhu, F., Yang, H., Yi, L., et~al.: Humanbench: Towards general human-centric perception with projector assisted pretraining. In: IEEE Conf. Comput. Vis. Pattern Recog. pp. 21970--21982 (2023)

\bibitem{tian2015pedestrian}
Tian, Y., Luo, P., Wang, X., Tang, X.: Pedestrian detection aided by deep learning semantic tasks. In: IEEE Conf. Comput. Vis. Pattern Recog. pp. 5079--5087 (2015)

\bibitem{tian2019directpose}
Tian, Z., Chen, H., Shen, C.: Directpose: Direct end-to-end multi-person pose estimation. arXiv preprint arXiv:1911.07451  (2019)

\bibitem{tian2020conditional}
Tian, Z., Shen, C., Chen, H.: Conditional convolutions for instance segmentation. In: Eur. Conf. Comput. Vis. pp. 282--298 (2020)

\bibitem{wang2022contextual}
Wang, D., Zhang, S.: Contextual instance decoupling for robust multi-person pose estimation. In: IEEE Conf. Comput. Vis. Pattern Recog. pp. 11060--11068 (2022)

\bibitem{wang2020deep}
Wang, J., Sun, K., Cheng, T., Jiang, B., Deng, C., Zhao, Y., Liu, D., Mu, Y., Tan, M., Wang, X., et~al.: Deep high-resolution representation learning for visual recognition. IEEE Trans. Pattern Anal. Mach. Intell.  (2020)

\bibitem{wang2020towards}
Wang, Z., Zheng, L., Liu, Y., Li, Y., Wang, S.: Towards real-time multi-object tracking. In: Eur. Conf. Comput. Vis. pp. 107--122 (2020)

\bibitem{wei2020point}
Wei, F., Sun, X., Li, H., Wang, J., Lin, S.: Point-set anchors for object detection, instance segmentation and pose estimation. In: Eur. Conf. Comput. Vis. pp. 527--544 (2020)

\bibitem{wojke2017simple}
Wojke, N., Bewley, A., Paulus, D.: Simple online and realtime tracking with a deep association metric. In: IEEE Int. Conf. Image Process. pp. 3645--3649 (2017)

\bibitem{xiao2018simple}
Xiao, B., Wu, H., Wei, Y.: Simple baselines for human pose estimation and tracking. In: Eur. Conf. Comput. Vis. (2018)

\bibitem{xie2020polarmask}
Xie, E., Sun, P., Song, X., Wang, W., Liu, X., Liang, D., Shen, C., Luo, P.: Polarmask: Single shot instance segmentation with polar representation. In: IEEE Conf. Comput. Vis. Pattern Recog. pp. 12193--12202 (2020)

\bibitem{xu2021vipnas}
Xu, L., Guan, Y., Jin, S., Liu, W., Qian, C., Luo, P., Ouyang, W., Wang, X.: Vipnas: Efficient video pose estimation via neural architecture search. In: IEEE Conf. Comput. Vis. Pattern Recog. pp. 16072--16081 (2021)

\bibitem{xu2022zoomnas}
Xu, L., Jin, S., Liu, W., Qian, C., Ouyang, W., Luo, P., Wang, X.: Zoomnas: searching for whole-body human pose estimation in the wild. IEEE Trans. Pattern Anal. Mach. Intell.  \textbf{45}(4),  5296--5313 (2022)

\bibitem{xu2022vitpose}
Xu, Y., Zhang, J., Zhang, Q., Tao, D.: Vitpose: Simple vision transformer baselines for human pose estimation. Adv. Neural Inform. Process. Syst.  \textbf{35},  38571--38584 (2022)

\bibitem{xue2022learning}
Xue, N., Wu, T., Xia, G.S., Zhang, L.: Learning local-global contextual adaptation for multi-person pose estimation. In: IEEE Conf. Comput. Vis. Pattern Recog. pp. 13065--13074 (2022)

\bibitem{yang2023explicit}
Yang, J., Zeng, A., Liu, S., Li, F., Zhang, R., Zhang, L.: Explicit box detection unifies end-to-end multi-person pose estimation. Int. Conf. Learn. Represent.  (2023)

\bibitem{zeng2022not}
Zeng, W., Jin, S., Liu, W., Qian, C., Luo, P., Ouyang, W., Wang, X.: Not all tokens are equal: Human-centric visual analysis via token clustering transformer. In: IEEE Conf. Comput. Vis. Pattern Recog. pp. 11101--11111 (2022)

\bibitem{zeng20203d}
Zeng, W., Ouyang, W., Luo, P., Liu, W., Wang, X.: 3d human mesh regression with dense correspondence. In: IEEE Conf. Comput. Vis. Pattern Recog. pp. 7054--7063 (2020)

\bibitem{zhang2022dino}
Zhang, H., Li, F., Liu, S., Zhang, L., Su, H., Zhu, J., Ni, L.M., Shum, H.Y.: Dino: Detr with improved denoising anchor boxes for end-to-end object detection. Int. Conf. Learn. Represent.  (2023)

\bibitem{zhang2014panda}
Zhang, N., Paluri, M., Ranzato, M., Darrell, T., Bourdev, L.: Panda: Pose aligned networks for deep attribute modeling. In: IEEE Conf. Comput. Vis. Pattern Recog. pp. 1637--1644 (2014)

\bibitem{zhang2020mask}
Zhang, R., Tian, Z., Shen, C., You, M., Yan, Y.: Mask encoding for single shot instance segmentation. In: IEEE Conf. Comput. Vis. Pattern Recog. pp. 10226--10235 (2020)

\bibitem{zhang2017citypersons}
Zhang, S., Benenson, R., Schiele, B.: Citypersons: A diverse dataset for pedestrian detection. In: IEEE Conf. Comput. Vis. Pattern Recog. pp. 3213--3221 (2017)

\bibitem{zhang2019pose2seg}
Zhang, S.H., Li, R., Dong, X., Rosin, P., Cai, Z., Han, X., Yang, D., Huang, H., Hu, S.M.: Pose2seg: Detection free human instance segmentation. In: IEEE Conf. Comput. Vis. Pattern Recog. pp. 889--898 (2019)

\bibitem{zhang2021location}
Zhang, X., Ma, B., Chang, H., Shan, S., Chen, X.: Location sensitive network for human instance segmentation. IEEE Trans. Image Process.  \textbf{30},  7649--7662 (2021)

\bibitem{zhang2024when}
Zhang, Y., Zeng, W., Jin, S., Qian, C., Luo, P., Liu, W.: When pedestrian detection meets multi-modal learning: Generalist model and benchmark dataset. In: Eur. Conf. Comput. Vis. (2024)

\bibitem{zhang2021fairmot}
Zhang, Y., Wang, C., Wang, X., Zeng, W., Liu, W.: Fairmot: On the fairness of detection and re-identification in multiple object tracking. Int. J. Comput. Vis.  \textbf{129},  3069--3087 (2021)

\bibitem{zhang2017quantifying}
Zhang, Y., Liu, L., Li, C., Loy, C.C.: Quantifying facial age by posterior of age comparisons. In: Brit. Mach. Vis. Conf. (2017)

\bibitem{zheng2022progressive}
Zheng, A., Zhang, Y., Zhang, X., Qi, X., Sun, J.: Progressive end-to-end object detection in crowded scenes. In: Proceedings of the IEEE/CVF conference on computer vision and pattern recognition. pp. 857--866 (2022)

\bibitem{zheng2015scalable}
Zheng, L., Shen, L., Tian, L., Wang, S., Wang, J., Tian, Q.: Scalable person re-identification: A benchmark. In: Int. Conf. Comput. Vis. (2015)

\bibitem{zhou2019objects}
Zhou, X., Wang, D., Kr{\"a}henb{\"u}hl, P.: Objects as points. arXiv preprint arXiv:1904.07850  (2019)

\bibitem{zhu2020deformable}
Zhu, X., Su, W., Lu, L., Li, B., Wang, X., Dai, J.: Deformable detr: Deformable transformers for end-to-end object detection. Int. Conf. Learn. Represent.  (2021)

\end{thebibliography}
